%% file: main.tex
\documentclass[11pt]{article}

\usepackage[preprint]{acl}

\usepackage{times}
\usepackage{latexsym}

\usepackage[T1]{fontenc}

\usepackage[utf8]{inputenc}

\usepackage{microtype}

\usepackage{inconsolata}

\usepackage{graphicx}

\input{section/macro}

\title{All Changes May Have Invariant Principles: Improving Ever-Shifting Harmful Meme Detection via Design Concept Reproduction}

\author{Ziyou Jiang$^{1,2,3}$, Mingyang Li$^{1,2,3}$\thanks{Corresponding author.}, Junjie Wang$^{1,2,3}$, \textbf{Yuekai Huang}$^{1,2,3}$, \\ \textbf{Jie Huang}$^{1,2,3}$, \textbf{Zhiyuan Chang}$^{1,2,3}$, \textbf{Zhaoyang Li}$^{1,2,3}$ \and \textbf{Qing Wang}$^{1,2,3*}$\\
        $^{1}$State Key Laboratory of Complex System Modeling and Simulation Technology,  \\ 
        Beĳing, China $^{2}$Science and Technology on Integrated Information System Laboratory \\ Institute of Software Chinese Academy of Sciences, Beĳing, China \\
        $^{3}$University of Chinese Academy of Sciences \\
        \textit{\{ziyou2019, mingyang2017, junjie, yuekai2018, huangjie, zhiyuan2019,}\\ \textit{lizhaoyang2024, wq\}@iscas.ac.cn,}}

\begin{document}
\maketitle

\input{section/abstract}
\input{section/introduction}
\input{section/background}

\input{section/method}
\input{section/exp}
\input{section/result}
\input{section/rw}

\input{section/conclusion}

\bibliography{custom}

\appendix
\input{section/appendix}
\end{document}

%% file: section/macro.tex
\usepackage{times}
\usepackage{latexsym}
\usepackage{booktabs}

\usepackage[T1]{fontenc}
\usepackage{multirow}
\usepackage{colortbl}
\usepackage{pifont}
\usepackage[ruled,linesnumbered]{algorithm2e}

\usepackage{amsthm} 
\newtheorem{theorem}{Theorem}
\newtheorem{lemma}{Lemma}

\usepackage[utf8]{inputenc}
\usepackage{amsmath}

\usepackage{microtype}
\usepackage{hyperref}

\usepackage{inconsolata}
\usepackage{threeparttable}

\usepackage{graphicx}
\usepackage{enumitem}
\usepackage{tcolorbox}
\usepackage{amsmath}
\usepackage{amsfonts}
\usepackage{mathtools}

\newcommand{\tool}{\textsc{RepMD}}

%% file: section/abstract.tex
\begin{abstract}

Harmful memes are ever-shifting in the Internet communities, which are difficult to analyze due to their type-shifting and temporal-evolving nature. Although these memes are shifting, we find that different memes may share invariant principles, i.e., the underlying design concept of malicious users, which can help us analyze why these memes are harmful. In this paper, we propose {\tool}, an ever-shifting harmful meme detection method based on the design concept reproduction. We first refer to the attack tree to define the Design Concept Graph (DCG), which describes steps that people may take to design a harmful meme. Then, we derive the DCG from historical memes with design step reproduction and graph pruning. Finally, we use DCG to guide the Multimodal Large Language Model (MLLM) to detect harmful memes. The evaluation results show that {\tool} achieves the highest accuracy with 81.1\% and has slight accuracy decreases when generalized to type-shifting and temporal-evolving memes. Human evaluation shows that {\tool} can improve the efficiency of human discovery on harmful memes, with 15$\sim$30 seconds per meme.

\textcolor{red}{\textbf{Disclaimer:} This paper may contain content that is disturbing to some readers.}

\end{abstract}

%% file: section/introduction.tex
\section{Introduction}



Nowadays, memes have emerged as important cultural symbols on the Internet.
They use both visual and textual elements to convey the designer's opinion for organizations, regimes, and social events~\cite{DBLP:journals/corr/abs-2401-01523}.
However, malicious users may design harmful memes as a weapon to express their biased viewpoints~\cite{sharma2022detecting}, which may pose threats to society and other users.

\begin{figure}[t]
\centering
\vspace{0.2cm}
\includegraphics[width=\columnwidth]{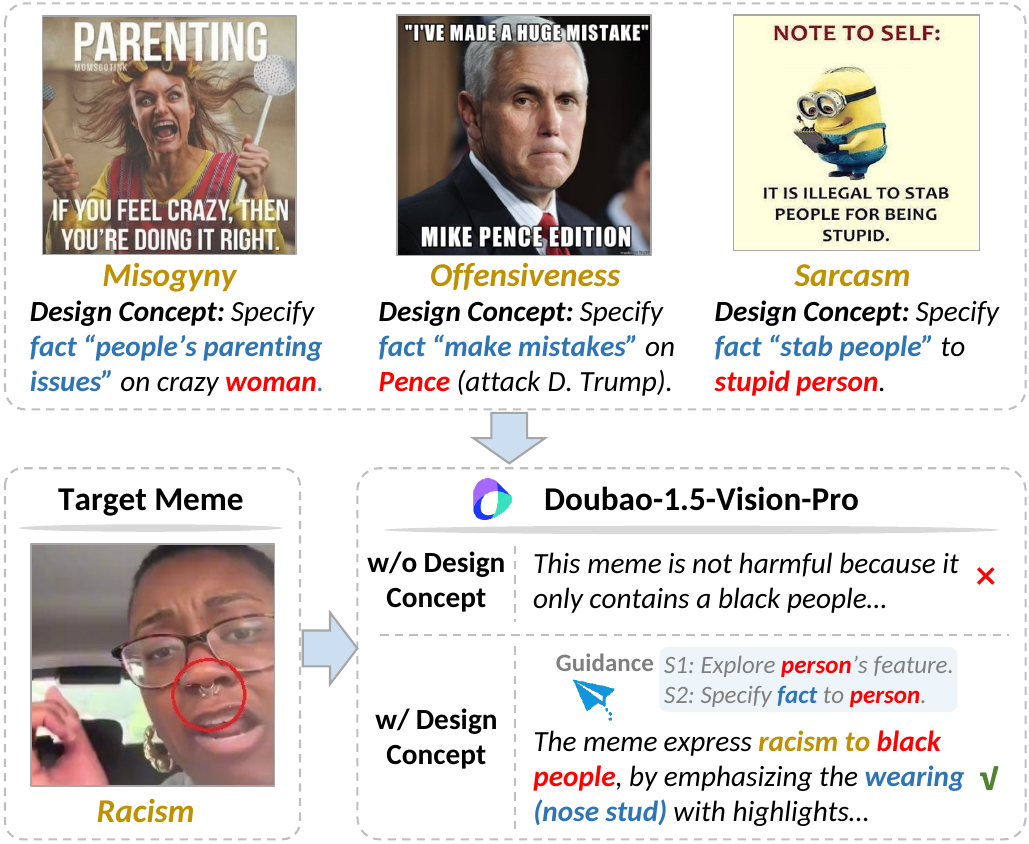}
\caption{The motivation example of {\tool}.}
\vspace{-0.3cm}
\label{fig:motivation}
\end{figure}

The rapid growth of the Internet community has led to the following ever-shifting characteristics of harmful memes~\cite{valensise2021entropy}:
\textbf{(1) Type Shifting:} Memes come in many forms and attack different targets, and \textbf{(2) Temporal Evolving:} Memes are temporally related to some events.
This ever-shifting nature makes the visual elements and expression of new memes quite different from the historical ones.
{For example, Figure \ref{fig:motivation}'s target meme uses a new and implicit way to express discrimination against black people, through the highlighted red circles on the people's accessories (i.e., nose stud).
Existing harmful meme detection methods~\cite{DBLP:conf/nips/KielaFMGSRT20, DBLP:conf/acl-trac/SuryawanshiCAB20} only learn the combination of harmful elements, and lack understanding of these implicit expressions, so they cannot accurately detect the harmfulness.
Moreover, new slang abbreviations like "\textit{GOAT}" and "\textit{Stan}" are used in memes to express harmful viewpoints, and these rarely-seen expression improve the difficulty of detection.}



Although harmful memes are ever-shifting, we find that they have \textbf{"invariant principles"}.
In Figure \ref{fig:motivation}, we choose four example memes that have different harmful types, but share similar underlying ideas that MLLM can learn from each other.
For example, the idea of the leftmost meme is to specialize in fact (e.g., parenting issues) to a certain group of people (e.g., crazy women), thereby attacking these people.
This is the \textbf{Design Concept} of memes, expressing the attacks through the \textit{human specification}~\cite{sualcudean2020visual}.
By analyzing the design concepts, the test MLLM, i.e., Doubao-1.5-Vision-Pro~\cite{DBLP:journals/corr/abs-2505-07062}, can accurately identify the target meme's racism: the malicious user specifies a fact "\textit{people who wear nose stud}" onto "\textit{black people}" to express the stereotype.
In the previous cases, we can see that the design concept is useful.
However, previous works have not defined an explainable structure to formally describe this design concept.
Therefore, we need to define its structure and derive the contents based on the visible information of memes.
Moreover, we also need to propose the corresponding usage strategy of design concepts, which can effectively guide the MLLM to detect target harmful memes.

In this paper, we propose an automated approach that \textbf{Rep}roduce memes' design concept to improve the ever-shifting harmful \textbf{M}eme \textbf{D}etection, named \textbf{{\tool}}. 
Inspired by the effectiveness of the attack tree~\cite{schneier1999attack}, we define the structure of DCG, a {\textbf{heterogeneous graph}}~\cite{DBLP:conf/kdd/ZhangSHSC19,gao2025bootstrapping} that describes steps that people may take to design a harmful meme, as well as the goal they want to achieve.
This structure can explain the user's design logic and guide MLLM to stepwise identify the harmfulness.
Based on this structure, we derived a DCG from historical memes that MLLMs fail to predict accurately, through the design steps' reproduction and graph pruning.
For target memes, we retrieve similar reproduction steps from DCG, then form the stepwise guidance to help MLLM detect target harmful memes in the ever-shifting scenario.

To evaluate the performance of {\tool}, 
{we conduct two types of experiments on a dataset of 58,192 memes.
For the type-shifting experiment on public GOAT-Bench memes, {\tool} achieves the highest accuracy with 81.1\% and has only a 2.1\% accuracy decrease in out-of-domain evaluation.
For the temporal-evolving experiment on our manually crawled Twitter memes, {\tool} also outperforms baselines and has a 0.3\% accuracy improvement in other quarters' evaluation.}
Moreover, our human evaluation shows that {\tool} can improve the efficiency of human discovery on harmful memes, with 15$\sim$30 seconds per meme.

This paper makes the following contributions:

\begin{itemize}[leftmargin=*]
    \item We propose {\tool}, a harmful meme detection method with design concept reproduction, applicable for ever-shifting memes on the Internet.
    \item We evaluate {\tool} on ever-shifting memes from two data sources, which outperforms baselines and can be generalized to type-shifting and temporal-evolving memes with the help of DCG.
    \item We conduct a human evaluation to illustrate that DCG has high explainability and can help evaluators manually identify harmful memes.
    \item We release the code and dataset\footnote{\url{https://github.com/jzySaber1996/RepMD}} to facilitate {\tool}'s reproducibility.
\end{itemize}

%% file: section/background.tex

\section{Definition of Design Concept}\label{sec:definition}

{To describe the meme's design concept, we refer to the idea of attack trees~\cite{schneier1999attack}, a threat model that contains STRIDE threat types, attack methods, goals, and logic of the attack reproduction.
It can guide LLM-based security testing tools to warn of software vulnerabilities~\cite{DBLP:journals/compsec/XiongL19}, which aligns with MLLM's logic for identifying memes' harmful information.}

Before extracting the DCG, we first analyze the reasons why harmful memes cannot be detected by MLLM, then form the fail reason tree.
It contains the type and fail reason nodes, as well as the type and link edges.
Then, we derive the fail reason node to the steps that the harmful users will take to design harmful memes, 
which contain the type, reproduction
method, goal, and logic gate nodes, as well as the type, link, and achievement edges.

\begin{figure*}[t]
\centering
\includegraphics[width=\textwidth]{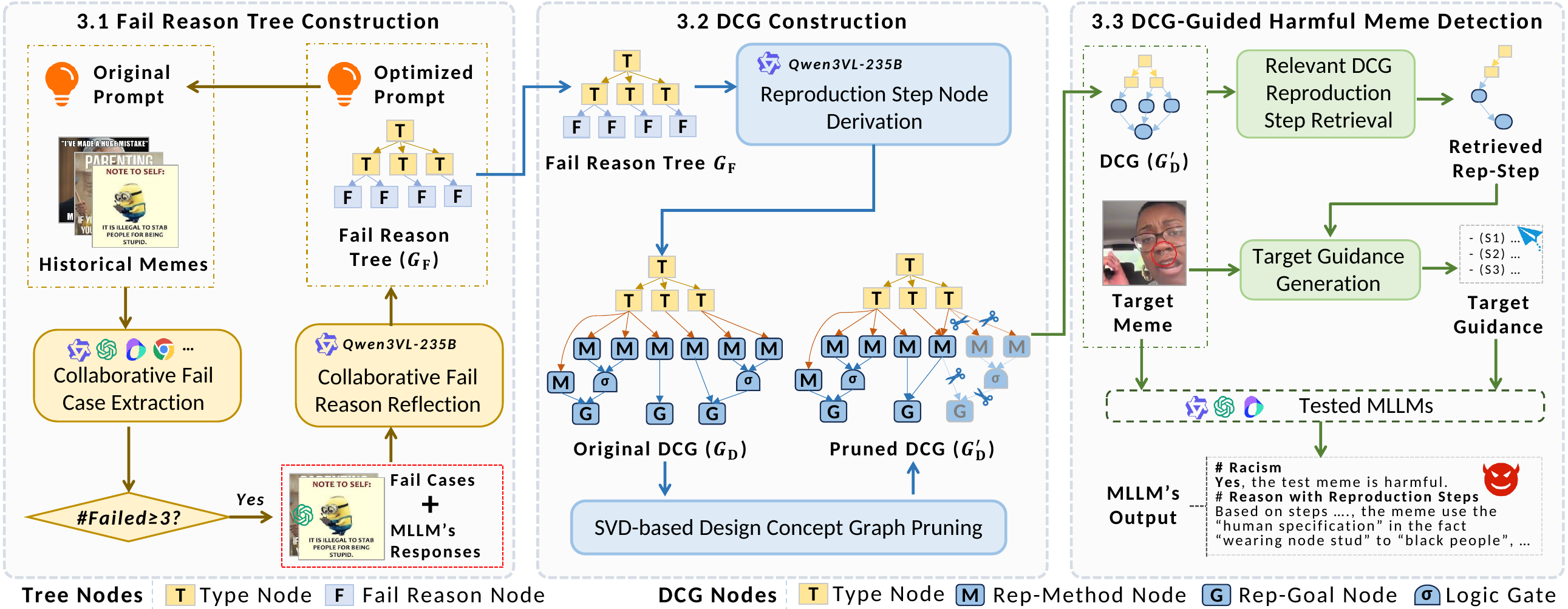}
\caption{The overview of {\tool}.}
\label{fig:model}
\end{figure*}

\subsection{Definition of Fail Reason Tree}
The structure of fail reason tree can be formed as $\mathcal{G}_\text{F}=\langle\{\mathcal{N}_\text{T},\mathcal{N}_\text{F}\},\{\mathcal{E}_\text{T},\mathcal{E}_\text{Link}\}\rangle$, defined as follows:

\noindent\textbf{Node:}
\textbf{(1) Type Node $\mathcal{N}_\text{T}$:} It contains three node levels $L_1\sim L_3$, where the chosen value is in Table \ref{tab:type_mapping},
similar to STRIDE threat types in attack trees:
\textbf{Macro Type Nodes $\mathcal{N}^{L_1}_\text{T}$:} The seven fixed categories in Table \ref{tab:type_mapping}, 
defined in the previous works~\cite{chen2025adammeme};
\textbf{Subtype Nodes $\mathcal{N}^{L_{2,3}}_\text{T}$:} The detailed types that need to be specified, e.g., \textit{Culture}$\rightarrow$\textit{Video Games}.
These subtypes are \textbf{not limited to the table} and can be extended.
\textbf{(2) Fail Reason Node $\mathcal{N}_\text{F}$:}
It contains {one-sentence MLLM's fail reason} and {memes' text description}.

\noindent\textbf{Edge:}
\textbf{(1) Type Edge $\mathcal{E}_\text{T}$:} The edge is $\mathcal{N}^{L_i}_\text{T}\rightarrow\mathcal{N}^{L_{i+1}}$, indicating the division from the macro type to subtypes;
\textbf{(2) Link Edge $\mathcal{E}_\text{Link}=\mathcal{N}_\text{T}\rightarrow\mathcal{N}_\text{F}$:} It indicates that one type of incorrectly predicted harmful meme has corresponding fail reasons.

\subsection{Definition of DCG}
We extract design concepts of historical memes and gather them into the DCG.
The structure refers to the attack tree's three-level structure, i.e., attack method, attack goal, and logic gate, which reflect how to reproduce a harmful meme.
The structure of DCG can be formed as $\mathcal{G}_\text{D}=\langle\{\mathcal{N}_\text{T},\mathcal{N}_\text{M},\mathcal{N}_\text{G},\mathcal{N}_\sigma\},\{\mathcal{E}_\text{T},\mathcal{E}_\text{A},\mathcal{E}_\text{Link}\}\rangle$, where the reproduction nodes are derived from reason nodes $\mathcal{N}_\text{F}\rightarrow(\mathcal{N}_\text{M},\mathcal{N}_\text{G},\mathcal{N}_\sigma)$, which is the reproduction step for the DCG. defined as follows:

\noindent\textbf{Node:} 
\textbf{(1) Type Node $\mathcal{N}_\text{T}$:} 
They are the same as the fail meme tree.
\textbf{(2) Reproduction Method $\mathcal{N}_\text{M}$:} These nodes describe the steps that malicious users take to design harmful memes;
\textbf{(3) Logic Gate $\mathcal{N}_{\sigma}$:} There is a logical combination between the reproduction methods to design a meme, i.e., \textit{And} ($\land$), \textit{Or} ($\lor$), and \textit{Not} ($\lnot$);
\textbf{(4) Reproduction Goal $\mathcal{N}_\text{G}$:} A design goal that a malicious user aims to achieve, such as the \textit{{human specification}}.
Each node also contains a harmful indicator $\{0,1\}$ that identifies which node may pose a harmful.

\noindent\textbf{Edge:}
\textbf{(1) Type Edge $\mathcal{E}_\text{T}$:} 
It is the same as the fail meme tree.
\textbf{(2) Link Edge $\mathcal{E}_\text{Link}=\mathcal{N}_\text{T}\rightarrow\mathcal{N}_\text{M}$:}
The structure is the same as the fail meme tree, but the meaning is slightly different. The link indicates that in historical memes, one type of harmful meme has the corresponding reproduction step.
\textbf{(3) Achievement Edge $\mathcal{E}_\text{A}$:}
Edges between $(\mathcal{N}_\text{M},\mathcal{N}_\text{G},\mathcal{N}_\sigma)$, which are the logic indicating the achievement between methods and the goal.

%% file: section/method.tex
\section{Overview of {\tool}}\label{sec:overview}

The element changing in ever-shifting harmful memes makes them difficult to detect.
As shown in Figure \ref{fig:motivation}, the target meme expresses racism against people in an unknown and implicit way.
With the help of the design concepts, we find that different types of memes may have invariant principles, such as the above-mentioned \textit{human specification}.
This invariant feature can help MLLMs infer why the unknown meme is harmful.

Figure \ref{fig:model} illustrates the overview of {\tool}. 
Guided by the concept, we first explore the reasons in historical meme cases that MLLMs fail to detect harmful memes, formed as the fail reason tree defined in Section \ref{sec:definition}.
Second, we derive reproduction steps from the fail reason tree and form DCG, then propose a fast {Singular Value Decomposition (SVD)}-based graph pruning method, proved effective in GNN's dimensionality reduction~\cite{DBLP:conf/iclr/Cai0XR23}.
These steps form the design concept with the relation "\textit{\textbf{Fail Reason Tree$\rightarrow$DCG$\rightarrow$SVD}}".
Finally, {\tool} retrieves relevant reproduction steps from DCG to help MLLMs detect whether the target meme is harmful.

\subsection{Fail Reason Tree Construction}
\label{sec:failed}
Before constructing the DCG $\mathcal{G}_\text{D}$, we aim to explore the reason why MLLMs fail to detect harmful memes.
Given historical memes $[M^{H}_1,...,M^{H}_n]$, we propose an looped framework of "\textit{extraction, reflection, and optimization}". 
After the loop, we can obtain a fail reason tree $\mathcal{G}_\text{F}$ and the optimized prompt. 
Through prompt optimization, the fail reason tree will contain some harmful memes that cannot be accurately detected, no matter how the prompt is optimized.  Therefore, it is meaningful to explore design concept for these memes.



\subsubsection{{Collaborative Fail Case Extraction}}

For each historical meme $M^{H}_{i}$, given the original $P_\text{Harm}$ (see Figure \ref{fig:detection_prompt}), we input it into the five tested MLLMs, i.e., Doubao-1.5-Vision, GPT-4o, Qwen2.5VL, Gemini-Pro, and InternVL3, based on evaluation reports~\cite{DBLP:journals/corr/abs-2401-15071,li2024seed}. 
These MLLMs achieve state-of-the-art performances in tasks like OCR and Image Understanding, etc., with high \textbf{Performance} and \textbf{Time Efficiency}.
Some novel models, like Qwen3VL-235B, may take over two minutes with a long chain-of-thought, so we do not evaluate them.
We then obtain fail cases $M_\text{fail}$ with the major voting, i.e., \text{$\geq3$ MLLMs Fail}, which means in the $i_{th}$ iteration, even prompts are optimized, MLLMs still struggle to identify their harmfulness, so they need to be prioritized to analyze.

\subsubsection{Collaborative Fail Reason Reflection}
Given fail cases $M_\text{fail}$, we use the tree generation MLLM with prompt $P_\text{F}$ (see Figure \ref{fig:prompt_tree_construction}) to generate $\mathcal{G}_\text{F}$:
\textbf{(1) Reason Nodes Generation:}
We concatenate the responses of all fail MLLMs, then ask a larger MLLM (i.e., Qwen3VL-235B) to analyze why they cannot detect the harmful: $M_\text{fail}\rightarrow\mathcal{N}_\text{F}$.
We directly tell Qwen3VL that MLLMs fail in prompt $P_\text{F}$ and ask it to generate content in the fail reason nodes.
\textbf{(2) Type Nodes Generation:}
To analyze which type the meme belongs to, we input the pre-defined types and the reason node's content into Qwen3VL and categorize it into type nodes:
$(M_\text{fail},\mathcal{N}_\text{F})\rightarrow\mathcal{N}_\text{T}$. 
The prompt controls MLLM to first classify the meme into the macro type as $L_1$,
then stratify subtypes from $L_2$ to $L_3$. 
\textbf{(3) Prompt Optimization.} 
We use Qwen3VL to summarize the key points in $\mathcal{G}_\text{F}$, which summarize the common causes of fail memes and what the MLLM needs to focus on.
We optimize the prompt by appending these points: $P_\text{Harm}\oplus \text{KeyPoints}\rightarrow P^{'}_\text{Harm}$, which makes MLLMs focus on fail cases and pertinently improve the performance of meme detection.

\subsection{DCG Construction}\label{sec:dcg}

\begin{figure}[t]
\centering
\includegraphics[width=\columnwidth]{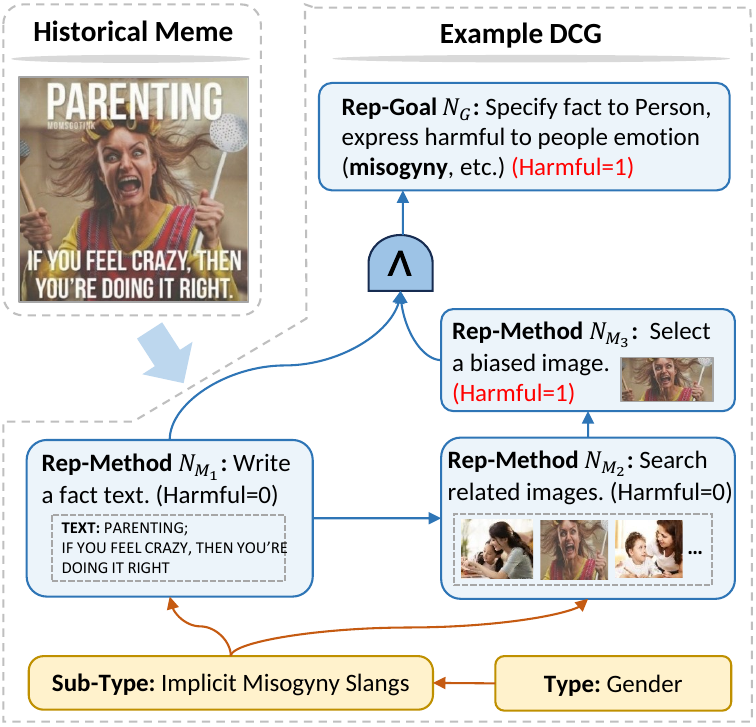}
\vspace{-0.5cm}
\caption{The example DCG extracted from the historical meme (harmful type: misogyny).}
\label{fig:example_dcg}
\vspace{-0.4cm}
\end{figure}

Given the fail reason tree $\mathcal{G}_\text{F}$ with the "reason-level expression" (e.g., Figure \ref{fig:example_dcg}'s "\textit{crazy woman with the parenting issues}"), we aim to explore memes' "reproduction-level idea" (e.g., "\textit{deliberately choose a crazy woman to emphasize and attacking these persons}).
The "reproduction" means thinking about the user's design process based on visible information.
Its core idea is to find a harmless alternative element, thereby identifying the reason why users choose the current element.

We propose the node derivation and calibration method to derive the reproduction steps and form the original DCG: $\mathcal{G}_\text{D}$.
Then, since $\mathcal{G}_\text{D}$ may contain redundant nodes, and MLLMs may have excessive time cost to prune this graph, we propose an effective pruning algorithm based on the SVD.

\subsubsection{Reproduction Step Node Derivation}
For reason nodes $\mathcal{N}_\text{F}$, we derive reproduction steps with Qwen3VL-235B and prompt $P_\text{D}$ (see Figure \ref{fig:prompt_dcg_construction}): $\mathcal{N}_\text{F}\rightarrow \mathcal{G}_\text{D}$:
\textbf{(1) Reproduction Method and Logic Gate Derivation:} We ask Qwen3VL to capture the reason-level elements with the combination logic $\mathcal{N}_\sigma$. 
Then, according to the previous description of alternative design approaches, we drill down into the ideas and steps that focus on the method of designing this meme $\mathcal{N}_\text{M}$, by asking Qwen3VL three questions:
"\textit{\textbf{A:} Is there a replacement method for each element?}", "\textit{\textbf{B:} Why is that element chosen?}", and "\textit{\textbf{C:} Is the replaced element harmful?}".
\textbf{(2) Graph Calibration and Goal Derivation:}
We ask another Qwen3VL to validate whether the design concept aligns with the meme's visual elements and analyze what the user wants to achieve as $\mathcal{N}_\text{G}$.
After generating design concepts from historical memes, we link them by type nodes and form an overall graph as DCG.

\subsubsection{SVD-based DCG Pruning}\label{sec:retried}
The DCG is the gathering of all historical memes, so some memes may have similar design concepts, introducing redundant nodes and edges to the DCG. 
Therefore, we propose the SVD-based pruning method, as shown in Algorithm \ref{alg:pruning}.
The main idea is how to design an \textbf{adjacency matrix} to describe the relations between nodes $\mathcal{N}_i$ and $\mathcal{N}_j$, so we propose the reproduction score (\textbf{line \ref{algo_line:rep_score}}): 
\begin{equation}\label{equa:rep_score}
\resizebox{.89\linewidth}{!}{$
    \displaystyle  
    \begin{split}
        &Score_\text{rep}(\mathcal{N}_{ij})=\text{ReLU}(sim(\mathcal{N}_i, \mathcal{N}_{root})-sim(\mathcal{N}_i, \mathcal{N}_{{j}}))\\
        &\text{where }sim(\mathcal{N}_i,\mathcal{N}_j)=0\text{ if }\mathcal{N}_i/\mathcal{N}_j=\mathcal{N}_\text{root}
    \end{split} 
$}
\end{equation}
where the function $sim(\cdot,\cdot)$ is the cosine similarity between TF-IDF values of the node's contents. 
$\mathcal{N}_\text{T}$'s root node is $\mathcal{N}^{L_1}_\text{T}$, and $\mathcal{N}_\text{M}$'s root node is $\mathcal{N}_{G}$. The final scores are activated with the ReLU function~\cite{glorot2011deep} to ensure the positive definiteness of the adjacent matrix.
The equation implies that we want a node to have \textbf{higher correlation} to the root node and \textbf{lower similarity} to other nodes, which means that this node is not redundant and will not be pruned.
Moreover, we find that redundancy mainly comes from reproduction steps, so we introduce scaled factors $\alpha$, $\beta$ for the matrix of the reproduction step $(\mathcal{N}_\text{M},\mathcal{N}_\text{G},\mathcal{N}_\sigma)\rightarrow\mathbf{A}_\text{Repr}$ and the edge matrix $\mathcal{E}_\text{Link}\rightarrow\mathbf{A}_\text{Link}$ ($1>\beta>\alpha$).

From \textbf{line \ref{algo_line:start_loop} to \ref{algo_line:end_loop}}, after we have initialized the base matrix $\mathbf{A}$, we calculate the 
$t$-hop matrix's indirect feature value, i.e., $\mathbf{A}^{t}$ in the loop, which indicates how the nodes are correlated to other nodes through the graph's relations.
Then, we set a hyperparameter $\theta$ to represent the retained nodes in $\mathcal{G}^{'}_\text{D}$.
We first introduce the Laplacian Normalization~\cite{li2024beyond} to process the adjacent matrix to $\mathbf{L}$. 

In \textbf{line \ref{algo_line:svd}}, during the SVD, we obtain singular values $[\lambda_1,...\lambda_n]$, which can be used to represent the matrix's feature with the low-rank approximation.
Since the distribution of singular values may be long-tailed, we calculate the difference on a logarithmic scale and choose the cut-off value $\lambda_{cut}$ with the steepest decline.

Finally, in \textbf{line \ref{alg:add_t_hop_relation}}, if the proportion of retained nodes does not achieve $\theta$, we explore the impact of the $t$-hop matrix's indirect relations.

\input{algo/pruning}

\subsection{DCG-Guided Harmful Meme Detection}\label{sec:retried}

Since {\tool} has constructed DCG with multiple prompts, the detection is lightweight by reusing previous prompts.
Given target memes $[M_1^{T},...,M_n^{T}]$ and the test MLLM, we retrieve the similar reproduction steps from the DCG $\mathcal{G}^{'}_{R}$ and guide the harmful meme detection.
Since this step is executed online, the time cost of {\tool} comes from the scale of the test dataset.

\paragraph{Step-1: Relevant DCG Reproduction Step Retrieval.}
We extract the historical reproduction step as follows: \textbf{(1) DCG Node Retrieval:} We reuse the prompt $P_\text{F}$ and $P_\text{D}$ to select DCG's nodes that may relate to $M^T_i$, aggregated as a set $\{\mathcal{N}_i|\mathcal{N}_i\in\mathcal{G}^{'}_\text{D}\}$.
\textbf{(2) Reproduction Step Formation:} We cluster the subgraphs based on the edges, i.e., $\mathcal{G}_\text{sub}=\langle \{\mathcal{N}_i\},\mathcal{E}_\text{sub}\subset\mathcal{G}^{'}_\text{D}\rangle$.

\paragraph{Step-2: Target Guidance Generation.}
We transform the retrieved step $\mathcal{G}_\text{sub}$ into plain text with logic semantics (see Appendix \ref{app:dcg_structure}), then feed the target meme and this plain text into the test MLLM (same model for testing the follow-up harmful meme detection) to output the guidance as $(S_1)\rightarrow(S_n)$. 
Finally, {\tool} inputs the generated guidance with the target meme $M_i^{T}$ into the test MLLM, then uses the optimized prompt $P^{'}_\text{Harm}$ in Figure \ref{fig:detection_prompt} to detect the target harmful memes.

%% file: algo/pruning.tex
\begin{algorithm}[t]
\small
	\caption{SVD-based DCG pruning.} 
 \label{alg:pruning}
	\KwIn{The original DCG $\mathcal{G}_\text{D}$, retained proportion $\theta$.} 
	\KwOut{The pruned DCG $\mathcal{G}^{'}_\text{D}$.}
    \SetKwProg{Fn}{Function}{}{end}
    $\mathbf{A}={Score_\text{rep}}(\mathcal{G}_\text{D})$ with Equation (\ref{equa:rep_score}), where 
    $\mathbf{A}=\begin{pmatrix}
    \mathbf{A}_\text{Type}&\beta\mathbf{E}_\text{Link}\\
    \beta\mathbf{E}_\text{Link}&\alpha\mathbf{A}_\text{Repr}
    \end{pmatrix}$, $\alpha,\beta$: scaled factors\;
    Calculate Degree Matrix $\mathbf{D}$, Initialize $\mathcal{G}^{'}_\text{D}=\emptyset$\;\label{algo_line:rep_score}
    \While {$t\leq 5$}{\label{algo_line:start_loop}
        $\mathbf{L}=\mathbf{I}-\mathbf{D}^{-1/2}\mathbf{A}^{t}\mathbf{D}^{-1/2}$\;
        SVD: $\mathbf{L}=\mathbf{U}\mathbf{\Lambda}\mathbf{V}^{\top}$ (See Appendix \ref{app:proof-svd}); $\Lambda=\text{diag}(\lambda_1,...,\lambda_n), \lambda_1\geq\lambda_2\geq...\geq0$\;\label{algo_line:svd}
        Cut-off Determination (See Appendix \ref{app:proof-cutoff}): $cut=\arg\max_{i}|\ln(\lambda_{i+1}-\lambda_{i})|$\;
        Graph Pruning: $\mathcal{G}^{'}_\text{D}=\langle\{\mathcal{N}_1,...,\mathcal{N}_{cut}\},\{\mathcal{E}^{'}\}\rangle$, where $\mathcal{N}_i:=\lambda_i$, $\{\mathcal{E}^{'}\}= \{\mathcal{N}_i\rightarrow\mathcal{N}_j|i,j\leq cut\}$\;
        \If {$cut/n\geq \theta$}{
        \textbf{break}\;
        }
        $t$-hop Indirect Features: $t=t+1$\;\label{alg:add_t_hop_relation}
    }\label{algo_line:end_loop}
    return $\mathcal{G}^{'}_\text{D}$\;
\end{algorithm}

%% file: section/exp.tex
\section{Experimental Design}\label{sec:data_preparation}
To evaluate the performance of {\tool}, we introduce three Research Questions (RQs).

\textit{\textbf{RQ1:} What are the performances of {\tool} on detecting ever-shifting harmful memes, i.e., type-shifting or temporal-evolving memes.} 

\textit{\textbf{RQ2:} How do each component contribute to {\tool}'s performances?} 
    
\textit{\textbf{RQ3:} Can {\tool}'s DCG help humans understand why the memes are harmful?}    

\paragraph{Dataset Preparation.}
We construct the dataset from two sources, i.e., GOAT-Bench~\cite{DBLP:journals/corr/abs-2401-01523} and our crawled memes from Twitter, and the steps are shown as follows.

\textbf{(1) Type-Shifting Dataset Preprocessing:} We first reuse all memes in GOAT-Bench. Then, we manually find that it has overlap categories (misogyny and hatefulness) and misclassified samples (classify misogyny memes as hatefulness), so we replace the type name, e.g., "Harmfulness"$\rightarrow$"Toxicity" (violence and suicide, etc) and "Hatefulness"$\rightarrow$"Racism" (mainly the racial hatefulness). Then, we reclassify 110 misogyny memes to the correct type.
The revised data is used to detect type-shifting harmful memes.

\textbf{(2) Temporal-Evolving Dataset Annotation:}
There are currently no usable benchmarks, so we manually crawl the memes on Twitter in \textbf{year: 2025} and divide them into four quarters. In each quarter, we randomly select 500 memes for DCG construction and evaluation.
We have invited three researchers with $\geq 3$ years of experience in harmful meme detection (not included in the authors), and asked them to annotate the dataset \textbf{independently} based on the criteria for determining harmful content~\cite{DBLP:journals/osnm/PandianiSC25}.
Each meme is labeled by these three annotators, and they check whether the labels are accurate.
The average Cohen's Kappa~\cite{DBLP:journals/jss/PerezDMT20} value is 0.86, which achieves a high agreement on the labels.

\input{table/dataset}

\input{table/res_cross_type_v3}

\textbf{(3) Dataset Size and Labeling Cost:}
Table \ref{tab:dataset} shows that we collect 58,192 memes (54.0\% are harmful) in the dataset, where 56,192 come from GOAT-Bench for type-shifting evaluation, and 2,000 are crawled memes for temporal-evolving experiments.
Moreover, the manually-labeled dataset comes from GOAT-Bench's adjustment and Twitter's, so the cost of human labor is small.

\paragraph{Baselines.}
We select three novel and representative baselines.
One method utilizes Retrieval-Augmented Generation (RAG) for harmful meme detection, which meets the evaluation requirements of DCG construction and retrieval.
The other two methods introduce few-shot Supervised Fine-Tuning (SFT), which meet the experimental requirements on ever-shifting memes: 
\textbf{Mode-Hate}~\cite{10.1145/3589334.3648145} is a few-shot hateful meme detector fine-tuned with LoRA; 
\textbf{MIND}~\cite{DBLP:conf/acl/Liu0LWD25} is a zero-shot RAG-based framework with bidirectional insight-augmented inference; and \textbf{RA-HMD}~\cite{mei-etal-2025-robust} is a fine-tuned adaptation framework for hateful meme detection.
Moreover, we have also selected four representative MLLMs as baselines, i.e., \textbf{Qwen2.5VL-32B}~\cite{DBLP:journals/corr/abs-2502-13923}, \textbf{GPT-4o}, \textbf{Doubao-1.5-Vision-Lite}, and \textbf{Doubao-1.5-Vision-Pro}~\cite{DBLP:journals/corr/abs-2505-07062}. 

\paragraph{Metrics.}
To ensure a fair and comprehensive evaluation, we measure model performance using two commonly-used metrics: \textbf{Accuracy} and \textbf{F1-Score}, following the setting of {GOAT-Bench}.

\paragraph{Experimental Settings.} 
For implementation, we use MLLM's APIs and set $temperature=0$ to make the output fixed, $Top\_P$, and $max\_token$ as default values in the vanilla MLLM.
For hyperparameters, we set $\alpha=0.3$ and $\beta=0.6$ for SVD, and $\theta=75\%$ (see Appendix \ref{app:other_res}).
All experiments run on four GeForce RTX A6000 GPUs.

%% file: table/dataset.tex
\begin{table}[htbp]
\vspace{-0.3cm}
\caption{The statistics of our constructed dataset.}
\vspace{-0.2cm}
\resizebox{\columnwidth}{!}{
\begin{tabular}{l|lllll}
\toprule
\multirow{2}{*}{\textbf{Types}} & \multicolumn{5}{c}{\textbf{Type-Shifting Dataset from GOAT-Bench}}   \\

& {Racism}  & {Misogyny} & {Offensiveness} & {Sarcasm} & {Toxity}         \\
\midrule
DCG      & 8,310            & 10,190            & 7,000                  & 19,816           & 4,250                   \\
Target            & 2,080            & 920               & 743                    & 1,820            & 1,063                   \\
\midrule

\multirow{2}{*}{\textbf{Quarters}} & \multicolumn{4}{c}{\textbf{Temporal-Evolving Dataset from Twitter}} & \multicolumn{1}{|c}{\multirow{2}{*}{\textit{\textbf{Total}}}} \\

& {Jan$\sim$Apr} & {Apr$\sim$Jun}  & {Jul$\sim$Sep}       & {Oct$\sim$Dec} & \multicolumn{1}{|c}{} \\
\midrule
DCG      & 400              & 400               & 400                    & 400              & \multicolumn{1}{|c}{\textit{51,166}}         \\
Target            & 100             & 100              & 100                   & 100             & \multicolumn{1}{|c}{\textit{7,026}}    \\
\bottomrule
\end{tabular}}
\vspace{-0.3cm}
\label{tab:dataset}
\end{table}

%% file: table/res_cross_type_v3.tex
\begin{table*}[t]
\small
\caption{The performance of {\tool} on In-Domain (ID) and Out-Of-Domain (OOD) memes(\%).}
\vspace{-0.2cm}

\resizebox{\textwidth}{!}{
\begin{tabular}{l|llllllllll|ll}
\toprule
\textbf{Evaluation on Types} & \multicolumn{2}{c}{\textbf{Racism}}                                     & \multicolumn{2}{c}{\textbf{Misogyny}}                                   & \multicolumn{2}{c}{\textbf{Offensiveness}}                              & \multicolumn{2}{c}{\textbf{Sarcasm}}                                    & \multicolumn{2}{c|}{\textbf{Toxicity}}                                   & \multicolumn{2}{c}{\textbf{\textit{Average}}}                                    \\
Approaches                 & F1                                 & Acc.                               & F1                                 & Acc.                               & F1                                 & Acc.                               & F1                                 & Acc.                               & F1                                 & Acc.                               & F1                                 & Acc.                               \\
\midrule
\multicolumn{13}{l}{\textit{Few-Shot SFT and RAG-based Baselines for Harmful Meme Detection}}                                                                                                                                                                                                                                                                                                                                                                                                            \\
\midrule
\quad Mod-Hate$_\text{ID}$~\cite{10.1145/3589334.3648145}          & 58.3                               & 58.1                               & 61.9                               & 61.5                               & 63.0                               & 64.1                               & 61.3                               & 63.3                               & 71.2                               & 69.3                               & 63.1                               & 63.3                               \\
\quad Mod-Hate$_\text{OOD}$        & 36.6                               & 34.9                               & 43.7                               & 44.5                               & 40.2                               & 43.7                               & 42.1                               & 45.2                               & 59.3                               & 55.1                               & 44.4                               & 44.7                               \\

\quad MIND$_\text{ID}$~\cite{DBLP:conf/acl/Liu0LWD25}              & 69.1                               & 66.2                               & 69.2                               & 68.7                               & 63.7                               & 62.5                               & 64.0                               & 63.2                               & 71.6                               & 71.2                               & 67.5                               & 66.4                               \\
\quad MIND$_\text{OOD}$            & 47.4                               & 45.0                               & 49.5                               & 49.4                               & 42.1                               & 42.0                               & 45.9                               & 44.3                               & 55.3                               & 54.0                               & 48.0                               & 46.9                               \\

\quad RA-HMD$_\text{ID}$~\cite{mei-etal-2025-robust}            & 74.6                               & 73.4                               & 77.3                               & 76.8                               & 71.9                               & 70.1                               & 70.4                               & 71.3                               & 70.2                               & 69.5                               & 72.9                               & 72.2                               \\
\quad RA-HMD$_\text{OOD}$          & 57.2                               & 56.3                               & 50.2                               & 48.6                               & 48.0                               & 47.4                               & 45.2                               & 44.3                               & 64.5                               & 62.3                               & 55.0                               & 51.8                               \\
\arrayrulecolor{lightgray}\midrule


\rowcolor[HTML]{ECF4FF}\quad \textit{Average} $\Delta(\text{OOD}, \text{ID})$         & \textcolor[HTML]{003300}{$\downarrow$20.3} &	\textcolor[HTML]{003300}{$\downarrow$20.5}	& \textcolor[HTML]{003300}{$\downarrow$21.7} &	\textcolor[HTML]{003300}{$\downarrow$21.5}	& \textcolor[HTML]{003300}{$\downarrow$22.8} &	\textcolor[HTML]{003300}{$\downarrow$21.2} &	\textcolor[HTML]{003300}{$\downarrow$20.8} &	\textcolor[HTML]{003300}{$\downarrow$21.3} &	\textcolor[HTML]{003300}{$\downarrow$11.3} &	\textcolor[HTML]{003300}{$\downarrow$12.9} &	\textcolor[HTML]{003300}{$\downarrow$18.7} &	\textcolor[HTML]{003300}{$\downarrow$19.5} \\

\arrayrulecolor{black}\midrule
\multicolumn{13}{l}{\textit{Test MLLMs for Harmful Meme Detection}}                                                                                                                                                                                                                                                                                                                                                                                                                         \\
\midrule
Qwen2.5VL-32B (Vanilla$^{*}$)        & 72.6                               & 68.3                               & 65.2                               & 60.9                               & 63.2                               & 55.3                               & 71.0                               & 68.7                               & 62.3                               & 53.9                               & 66.9                               & 61.4                               \\
\quad+{\tool}$_\text{ID}$            & 78.2                               & 77.4                               & 74.9                               & 74.5                               & 71.0                               & 70.6                               & 80.4                               & 79.6                               & 79.1                               & 77.0                               & 76.7                               & 75.8                               \\
\quad+{\tool}$_\text{OOD}$          & 76.8                               & 75.4                               & 73.0                               & 72.6                               & 70.2                               & 70.1                               & 77.2                               & 77.2                               & 77.5                               & 75.8                               & 74.9                               & 74.2                               \\
\arrayrulecolor{lightgray}\midrule
GPT-4o (Vanilla)                 & 76.6                               & 75.2                               & 82.7                               & 82.4                               & 59.1                               & 59.1                               & 68.7                               & 66.3                               & 66.3                               & 64.2                               & 70.7                               & 69.4                               \\
\quad+{\tool}$_\text{ID}$            & \textbf{88.5}                               & \textbf{87.6}                               & \textbf{90.5}                               & \textbf{90.4}                               & 71.2                               & 71.1                               & \textbf{84.2}                               & \textbf{83.8}                               & 80.2                               & 78.5                               & \textbf{82.9}                               & \textbf{82.3}                               \\
\quad+{\tool}$_\text{OOD}$          & 87.1                               & 87.0                               & 88.7                               & 86.2                               & 70.4                               & 70.2                               & 80.5                               & 79.6                               & 77.6                               & 76.9                               & 80.9                               & 80.0                               \\
\arrayrulecolor{lightgray}\midrule
Doubao-1.5-Vision-Lite (Vanilla) & 70.5                               & 69.1                               & 67.3                               & 68.0                               & 66.4                               & 65.5                               & 68.6                               & 68.7                               & 70.0                               & 70.2                               & 68.6                               & 68.3                               \\
\quad+{\tool}$_\text{ID}$            & 77.4                               & 77.2                               & 74.5                               & 74.1                               & 76.3                               & 75.4                               & 75.0                               & 74.9                               & 78.6                               & 78.5                               & 76.4                               & 76.0                               \\
\quad+{\tool}$_\text{OOD}$          & 76.2                               & 75.5                               & 73.2                               & 73.7                               & 72.1                               & 73.1                               & 74.1                               & 72.6                               & 77.5                               & 76.2                               & 74.6                               & 74.2                               \\
\arrayrulecolor{lightgray}\midrule
Doubao-1.5-Vision-Pro (Vanilla)   & 72.4                               & 71.9                               & 73.4                               & 72.9                               & 67.4                               & 64.3                               & 71.3                               & 70.4                               & 72.0                               & 72.1                               & 71.1                               & 69.9                               \\
\quad+{\tool}$_\text{ID}$            & 81.5                               & 80.6                               & 82.2                               & 81.3                               & \textbf{80.2}                               & \textbf{81.3}                               & 81.2                               & 81.3                               & \textbf{83.0}                               & \textbf{82.8}                               & 81.3                               & 81.1                               \\
\quad+{\tool}$_\text{OOD}$          & 80.3                               & 79.4                               & 80.4                               & 79.3                               & 78.2                               & 77.6                               & 78.5                               & 78.4                               & 81.7                               & 80.9                               & 79.4                               & 78.7                               \\
\arrayrulecolor{lightgray}\midrule


\rowcolor[HTML]{ECF4FF}\quad \textit{Average} $\Delta(\text{ID}, \text{Vanilla})$       & \textcolor[HTML]{990066}{$\uparrow$7.7}	& \textcolor[HTML]{990066}{$\uparrow$9.0}	& \textcolor[HTML]{990066}{$\uparrow$7.6}	& \textcolor[HTML]{990066}{$\uparrow$7.9} &	\textcolor[HTML]{990066}{$\uparrow$9.4}	& \textcolor[HTML]{990066}{$\uparrow$12.8} &	\textcolor[HTML]{990066}{$\uparrow$9.2}	& \textcolor[HTML]{990066}{$\uparrow$9.8} &	\textcolor[HTML]{990066}{$\uparrow$11.7}	 & \textcolor[HTML]{990066}{$\uparrow$13.1} &	\textcolor[HTML]{990066}{$\uparrow$9.1} &	\textcolor[HTML]{990066}{$\uparrow$10.5}\\


\rowcolor[HTML]{ECF4FF}\quad \textit{Average} $\Delta(\text{OOD}, \text{ID})$         & \textcolor[HTML]{003300}{$\downarrow$1.3}	& \textcolor[HTML]{003300}{$\downarrow$1.4}	& \textcolor[HTML]{003300}{$\downarrow$1.7} &	\textcolor[HTML]{003300}{$\downarrow$2.1}	& \textcolor[HTML]{003300}{$\downarrow$2.0} &	\textcolor[HTML]{003300}{$\downarrow$1.9}	& \textcolor[HTML]{003300}{$\downarrow$2.6} &	\textcolor[HTML]{003300}{$\downarrow$3.0}	& \textcolor[HTML]{003300}{$\downarrow$1.7} &	\textcolor[HTML]{003300}{$\downarrow$1.7}	& \textcolor[HTML]{003300}{$\downarrow$1.9} &	\textcolor[HTML]{003300}{$\downarrow$2.1}
\\

\arrayrulecolor{black}\bottomrule
\end{tabular}}
 \begin{tablenotes}
        \footnotesize
        \item[*] $^{*}$ "Vanilla" means using original MLLMs with the optimized detection prompt $P^{'}_\text{Harm}$ to detect harmful memes. 
      \end{tablenotes}
\label{tab:type_evolving_evalution}
\end{table*}

%% file: section/result.tex
\section{Results}

\subsection{Overall Performance of {\tool} on Evolving Memes (RQ1)}


\paragraph{Type-Shifting Experiments.}
We conduct this experiment in the following types: 
\textbf{(1) In-Domain ({ID}) Evaluation:} We construct DCG and detect target memes' harmfulness within the same type.
\textbf{(2) Out-Of-Domain ({OOD}) Evaluation:} We introduce the cross-type analysis, which means that we choose one specific type of memes in GOAT-Bench as the target, and use \textbf{the other four types of memes} as historical to generate DCG.

Table \ref{tab:type_evolving_evalution} illustrates the results for type-shifting scenario, where $\Delta(\text{ID},\text{Vanilla})$ and $\Delta(\text{OOD},\text{ID})$ indicate the absolute improvement of {\tool} and the relative decrease when migrated to the OOD memes.
We can see that, the absolute detection performance of {\tool} achieves 81.1\% accuracy, outperforming all baseline approaches.
The improvement on vanilla MLLMs achieves +9.1\% (F1) and +10.5\% (Accuracy).
Besides, the OOD experiment shows that the decrease of {\tool}'s performance is smaller than baselines (which is -18.7\% F1), with -1.9\% (F1) and -2.1\% (Accuracy). 
By analyzing the inference of MLLM, we find that they detect harmful memes by analyzing what elements are harmful, but may ignore the implicit expressions (e.g., Figure \ref{fig:motivation}'s red circle) and seemingly harmless elements. 
From the experiment results, we can see that DCG guides MLLM to understand the idea of memes' design, enabling {\tool} to detect harmful memes across types.
Even if there are not too many training samples labeled with fine-grained types, {\tool} can achieve effective detection ability by $\{0,1\}$ harmful labels.
Moreover, we have conducted a pairwise T-test to measure the significance between MLLMs and {\tool}\footnote{$p<0.01$ means high significance.}.
We can see that, in Table \ref{tab:significant_testing}, {\tool} can significantly outperform the vanilla MLLMs, and the column "\textbf{OOD vs ID}" shows that migrating {\tool} to type-shifting memes will not reduce the performance significantly.

\input{table/significant_testing}

\paragraph{Temporal-Evolving Experiments.}
We conduct this experiment in the following types:
\textbf{(1) Temporal Fixed (TF) Evaluation:} We construct DCG and evaluate {\tool} on same quarter's memes.
\textbf{(2) Temporal Evolving (TE) Evaluation:} To ensure fairness of evaluation, we choose \textbf{quarter$_i$}'s cases to construct DCG and detect harmful memes in \textbf{quarter$_{i+1}$}'s cases.

\input{table/time_emerged_v2}

Table \ref{tab:temporal_evolving_evalution} shows the Top-2 MLLMs' performances on Twitter's temporal-evolving memes within the last three quarters. 
We can see that, {\tool} improves the vanilla MLLMs with +13.7\% (F1) and +14.3\% (Accuracy) on average.
Moreover, we calculate the difference $\Delta(\text{TE},\text{TF})$ and find that introducing historical memes can even improve the detection performance of memes, with +0.8\% (F1) and +0.3\% (Accuracy) on average.
This advantage comes from the consistency with the design concepts of the same organization.
Some subcultural groups (e.g., Bitcoin's Dogwifhat, Minecraft, and K-pop Fancam, etc.) have opinion leaders, and group members will utilize a similar expression pattern to design memes, so the continuity of these patterns can help us identify the causes of harmful memes accurately.
Finally, in Table \ref{tab:significant_testing}'s column "\textbf{TE vs TF}", we can see that even when migrated to temporal evolving memes, {\tool}'s performance reductions are not significant, which further indicates the generalizability of our approach.

\begin{figure}[t]
\centering
\includegraphics[width=\columnwidth]{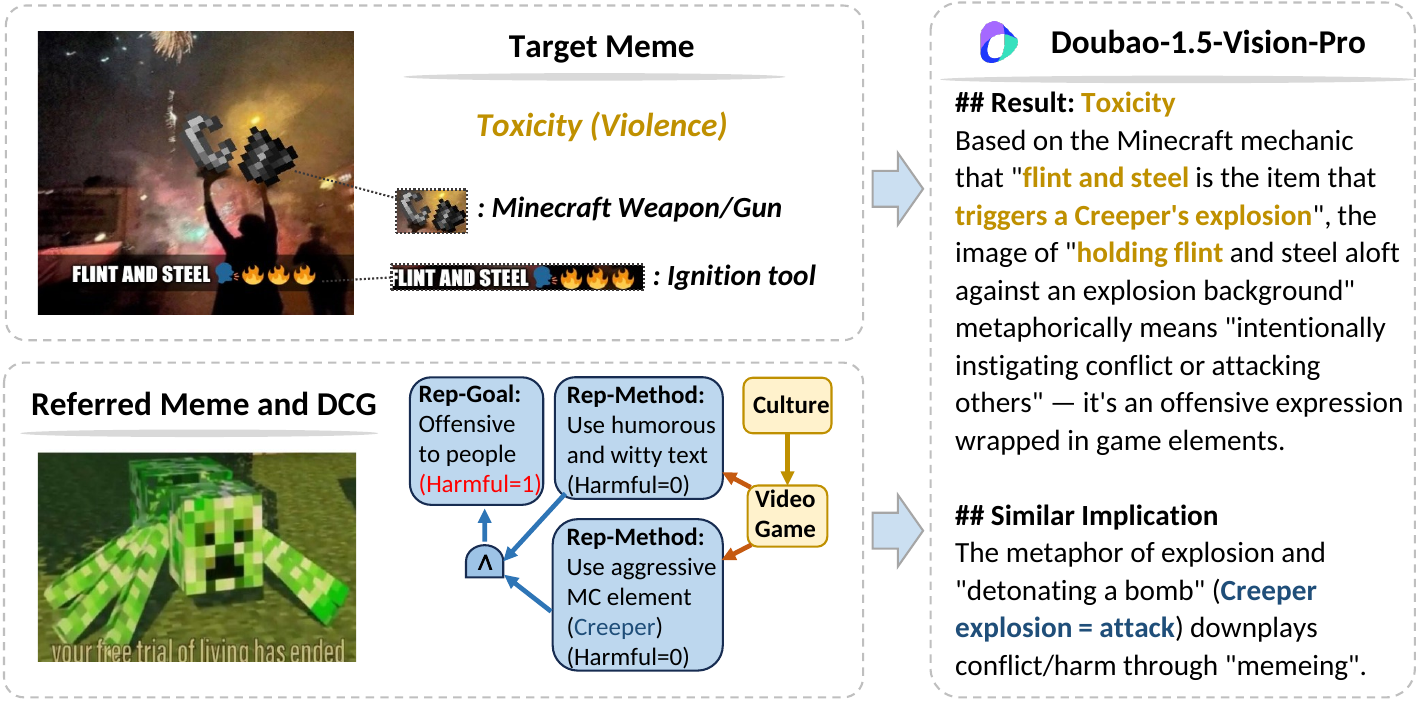}
\vspace{-0.3cm}
\caption{The case study of temporal evolving memes.}
\label{fig:case_study}
\end{figure}

\input{table/dcg_real_contribution}

\paragraph{Case Study for Design Concepts.}
Figure \ref{fig:case_study} shows that the Minecraft group's users use elements like "\textit{Creeper}" and "\textit{Ignition Tool}" instead of regular weapons to express violence, This expression is successfully captured by {\tool} in the DCG, so it can help the MLLMs understand ideas of the group's opinion leaders and members, thus detecting harmful memes that are rapidly shifting.

We have conducted another experiment to illustrate the advantages of GPT-4o+{\tool} with the help of DCG, where we manually select 200 examples with two design concepts, i.e., "\textit{Human Specification}" and "\textit{Minecraft Replacement}", where 150 memes are used for constructing DCG, and we detect harmful memes in the remaining 50 memes. The following results show the number of accurately predicted memes.
We can see that, in Table X, {\tool} can not only outperform baselines, but also detect 26 harmful memes with DCG, where vanilla GPT-4o cannot detect them.


\subsection{Ablation Study of {\tool} (RQ2)}

\input{table/ablation_removal_v2}

To evaluate the contribution of model's components to {\tool}, we analyze three types of \textbf{variants} based on the structure in Section \ref{sec:overview}: 
\textbf{Sec \ref{sec:failed}:}
Removing the failed reason tree and using all historical memes to construct DCG directly (\textbf{w/o Tree});
Removing the vote mechanism ({\textbf{w/o VoteMLLMs}}) and prompt optimization ({\textbf{w/o OptPrompt}});
\textbf{Sec \ref{sec:dcg}:}
Replacing DCG with fail reason tree ({\textbf{w/o SVD}}) and SVD with MLLMs ({\textbf{SVD$\rightarrow$MLLMs}});
\textbf{Sec \ref{sec:retried}:}
Directly generating reproduction steps without retrieval (\textbf{{w/o DCG}}), replacing retrieval with image similarity ({\textbf{Retrieval$\rightarrow$ImgSim}}), and the most similar graph-based RAG method, i.e., ({\textbf{Retrieval$\rightarrow$GraphRAG}}~\cite{peng2024graph}).

Table \ref{tab:ablation_study} shows the results of the ablation study.
We can see that, {\tool} outperforms most variants (5/6) in terms of the experiments ID, OOD, and TE. 
The maximum decrease is in the components that are related to the DCG, i.e., \textit{w/o SVD} and \textit{w/o DCG}, which achieve over -8.0\% and -9.0\% Accuracy.
Considering \textit{SVD$\rightarrow$MLLM}, we find that replacing SVD with Qwen3VL-235B fails on ID and OOD because of information truncation (too many input tokens) and incorrect output (pruned graph's nodes are different from original DCG). 
However, MLLM successfully prunes the graph and slightly outperforms SVD on the TE task, with a time cost of over 14$\times$ (see Appendix \ref{app:other_res}).
This indicates that SVD offers advantages in balancing time cost and performance, but fine-tuning an LLM-based pruning method may improve DCG quality and detection performance.

\subsection{Human Evaluation of {\tool} (RQ3)}

\begin{figure}[t]
\centering
\includegraphics[width=\columnwidth]{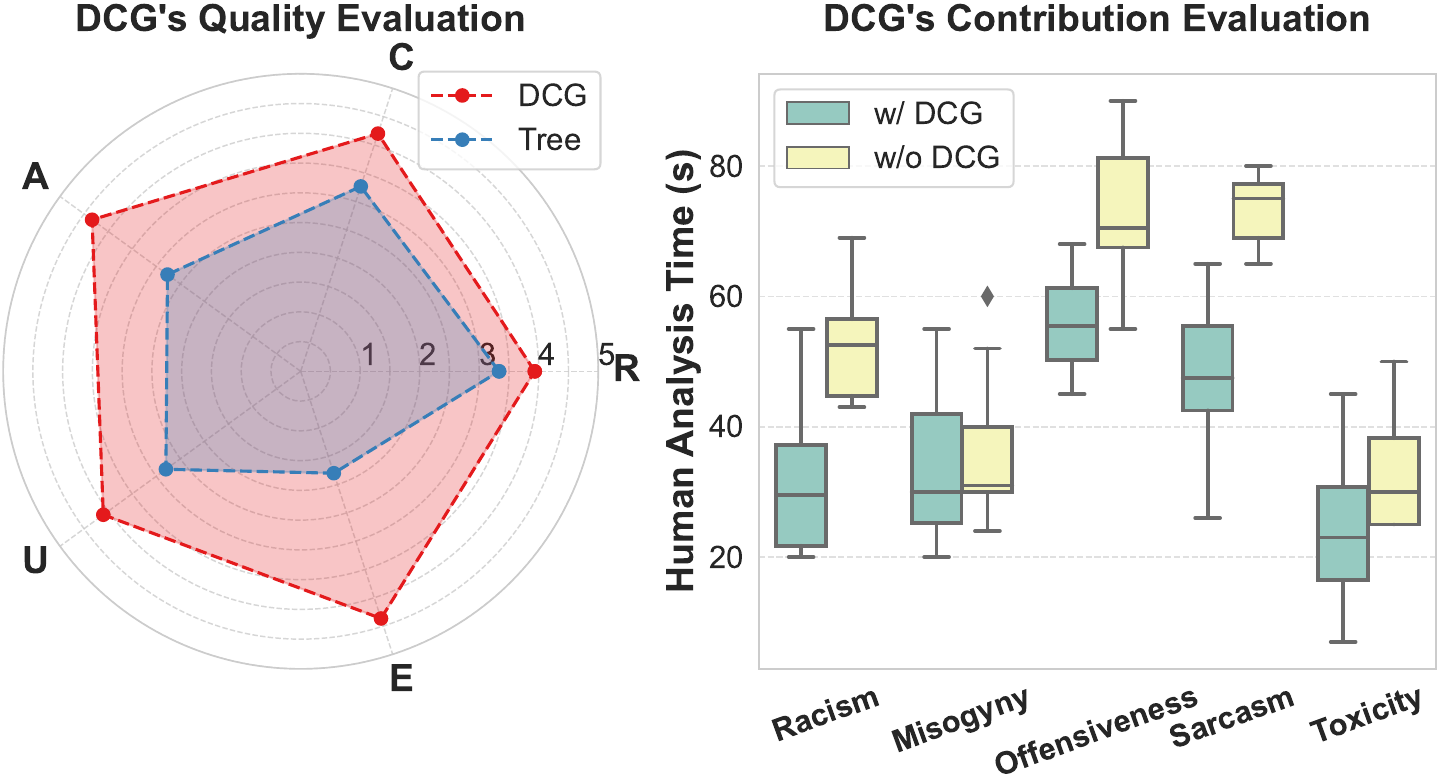}
\vspace{-0.6cm}
\caption{The results of human evaluation, i.e., scores of five evaluation criteria and analysis time cost.}
\label{fig:human_eval}
\end{figure}

We conduct the human evaluation on the DCG in two aspects: \textbf{(1) DCG's Quality:} We have invited three types of human evaluators categorized by research experiences on harmful memes (i.e., $\geq 3$ year veterans, $1\sim 3$ year beginners, and non-experienced social network users). Each category includes two evaluators.
We randomly select 10 historical memes with their corresponding fail reason trees and the DCG's design concepts, then rate these patterns on a 5-point Likert scale (from 1: poor to 5: excellent) based on five key criteria~\cite{mazhar2025figurative}: \textit{Relevance (R)}, \textit{Correctness (C)}, \textit{Actionability (A)}, \textit{Uniqueness (U)}, and \textit{Explainability (E, which is newly introduced based on DCG's structure)}.
\textbf{(2) DCG's Contribution:} We randomly select another 10 target harmful memes with generated target guidance, then calculate the time costs that previous evaluators take to identify their harmfulness with/without DCG's help. Note that we only count samples that can be successfully analyzed by humans. We have presented the questionnaire in Appendix \ref{app:human}.



Figure \ref{fig:human_eval} shows the results of human evaluation.
We can see that, for the overall DCG's quality evaluation, the DCG's five-dimensional scores outperform the fail reason trees.
All the scores are nearly or over 4, which means the DCG has high uniqueness and can be used to guide the detection of harmful memes.
Moreover, the explainability of the DCG ({score: 4.4}) outperforms that of the trees ({score: 1.8}) by the largest margin (score: +2.6), which means the DCG is easier to understand by evaluators with different experiences.
For the DCG's contribution, the time costs of human evaluation show that DCG can help evaluators improve their efficiency in accurately analyzing why memes are harmful.
Especially for the Offensiveness and Sarcasm memes that implicitly express the harmfulness, and humans cannot easily find their harmful elements, the DCG can reduce the average time costs from 15$\sim$30 seconds per meme.
Therefore, {\tool} can help humans identify harmful memes.
Experiments in Appendix \ref{app:text-to-image-safeguard} show that DCG can also help safeguard text-to-image models.

%% file: table/significant_testing.tex
\begin{table}[htbp]
\caption{The $p$-values of pairwise significant T-testing on type-shifting and temporal-evolving memes.}

\vspace{-0.2cm}
\resizebox{\columnwidth}{!}{
\begin{tabular}{l|lll}
\toprule
\textbf{Tested MLLMs}                   & \textbf{{\tool} vs Vanilla}   & \textbf{OOD vs ID} & \textbf{TE vs TF} \\
\midrule
Qwen2.5VL-32B          & $3.0\times 10^{-6}$  & 0.007     & 0.017    \\
GPT-4o                 & $7.0\times 10^{-9}$  & 0.037     & 0.032    \\
Doubao-1.5-Vision-Lite & $4.6\times 10^{-10}$ & 0.008     & 0.016    \\
Doubao-1.5-Vision-Pro  & $7.1\times 10^{-10}$ & 0.052     & 0.020  \\
\bottomrule
\end{tabular}}
\label{tab:significant_testing}
\end{table}

%% file: table/time_emerged_v2.tex
\begin{table}[htbp]
\small
\caption{The performance of {\tool} on Temporal Evolving (TE) and Temporal Fixed (TF) memes (\%).}
\vspace{-0.2cm}

\resizebox{\columnwidth}{!}{
\begin{tabular}{l|llllll}
\toprule
\textbf{Evaluation on Quarters} & \multicolumn{2}{c}{\textbf{Apr$\sim$Jun}$^{*}$}                        & \multicolumn{2}{c}{\textbf{Jul$\sim$Sep}}                              & \multicolumn{2}{c}{\textbf{Oct$\sim$Dec}}                               \\
Models (Top-2 MLLMs)   & F1                              & Acc.                            & F1                                 & Acc.                               & F1                                 & Acc.                               \\

\midrule

\quad RA-HMD$_\text{TF}$ & 45.9 & 44.9 & 52.7 & 51.5 & 56.4 & 55.2\\

\quad RA-HMD$_\text{TE}$ & 31.7 & 31.2 & 40.2 & 40.0 & 45.3 & 44.1\\

\rowcolor[HTML]{ECF4FF}\quad$\Delta(\text{TE}, \text{TF})$      & \textcolor[HTML]{003300}{$\downarrow$14.2}  & \textcolor[HTML]{003300}{$\downarrow$13.7} & \textcolor[HTML]{003300}{$\downarrow$12.5}     & \textcolor[HTML]{003300}{$\downarrow$11.5}    & \textcolor[HTML]{003300}{$\downarrow$11.1}    & \textcolor[HTML]{003300}{$\downarrow$11.1}    \\

\midrule
GPT-4o (Vanilla)                & 51.3                            & 50.0                            & 56.3                               & 55.0                               & 67.4                               & 65.0                               \\
\quad+{\tool}$_\text{TF}$                 & 60.2                            & 60.0                            & 63.0                               & 66.0                               & 83.5                               & 80.0                               \\
\quad+{\tool}$_\text{TE}$        & 67.4                            & 65.0                            & 61.6                               & 65.0                               & 82.4                               & 78.0                               \\
\rowcolor[HTML]{ECF4FF}\quad$\Delta(\text{TF}, \text{Vanilla})$      & \textcolor[HTML]{990066}{$\uparrow$8.9}  & \textcolor[HTML]{990066}{$\uparrow$10.0} & \textcolor[HTML]{990066}{$\uparrow$6.7}     & \textcolor[HTML]{990066}{$\uparrow$11.0}    & \textcolor[HTML]{990066}{$\uparrow$16.1}    & \textcolor[HTML]{990066}{$\uparrow$15.0}    \\
\rowcolor[HTML]{ECF4FF}\quad$\Delta(\text{TE}, \text{TF})$       & \textcolor[HTML]{990066}{$\uparrow$7.2}  & \textcolor[HTML]{990066}{$\uparrow$5.0}  & \textcolor[HTML]{003300}{$\downarrow$1.4} & \textcolor[HTML]{003300}{$\downarrow$1.0} & \textcolor[HTML]{003300}{$\downarrow$1.1} & \textcolor[HTML]{003300}{$\downarrow$2.0} \\
\arrayrulecolor{lightgray}\midrule
Doubao-1.5-V-Pro (Vanilla) & 67.4                            & 65.0                            & 66.2                               & 65.0                               & 67.4                               & 65.0                               \\
\quad+{\tool}$_\text{TF}$                 & 82.4                            & 80.0                            & 82.4                               & 80.0                               & 86.7                               & 85.0                               \\
\quad+{\tool}$_\text{TE}$        & 82.4                            & 80.0                            & 82.4                               & 80.0                               & 86.7                               & 85.0                               \\
\rowcolor[HTML]{ECF4FF}\quad$\Delta(\text{TF}, \text{Vanilla})$       & \textcolor[HTML]{990066}{$\uparrow$15.0} & \textcolor[HTML]{990066}{$\uparrow$15.0} & \textcolor[HTML]{990066}{$\uparrow$16.2}    & \textcolor[HTML]{990066}{$\uparrow$15.0}    & \textcolor[HTML]{990066}{$\uparrow$19.3}    & \textcolor[HTML]{990066}{$\uparrow$20.0}    \\
\rowcolor[HTML]{ECF4FF}\quad$\Delta(\text{TE}, \text{TF})$        & 0.0                             & 0.0                             & 0.0                                & 0.0                                & 0.0                                & 0.0                               
  \\
\arrayrulecolor{black}\bottomrule
\end{tabular}}
\begin{tablenotes}
        \footnotesize
        \item[*] $^{*}$ The \textit{\textbf{year: 2025}} is divided into quarter-based intervals. 
      \end{tablenotes}

\label{tab:temporal_evolving_evalution}
\end{table}

%% file: table/dcg_real_contribution.tex
\begin{table}[t]
\caption{The contribution of {\tool} on detecting harmful memes with different design concepts.}

\vspace{-0.2cm}
\resizebox{\columnwidth}{!}{
\begin{tabular}{l|ll}
\toprule
\textbf{Approach}          & \textbf{Human Specification} & \textbf{Minecraft Replacement} \\
\midrule
MIND           & 21/50 (42.0\%)       & 20/50 (40.0\%)         \\
RA-HMD         & 35/50 (79.0\%)       & 36/50 (72.0\%)         \\
Vanilla GPT-4o & 34/50 (68.0\%)       & 31/50 (62.0\%)         \\
\midrule
GPT-4o+{\tool}   & 44/50 (88.0\%)       & 47/50 (94.0\%)    \\
\bottomrule
\end{tabular}}
\label{tab:dcg_real_contribution}
\end{table}

%% file: table/ablation_removal_v2.tex
\begin{table}[t]
\small
\caption{The accuracy of {\tool}'s ablation study (component removal or replacement) on three variants (\%).}
\vspace{-0.2cm}

\resizebox{\columnwidth}{!}{
\begin{tabular}{l|lll}
\toprule
\textbf{Variants} & \textbf{ID}                             & \textbf{OOD}                            & \textbf{TE}                          \\
\midrule
GPT-4o+{\tool}      & 82.3                                      & 80.0                                      & 69.3                                      \\
\arrayrulecolor{lightgray}\midrule

{w/o Tree} & 81.1 \textcolor[HTML]{003300}{($\downarrow$1.2} & 79.2 \textcolor[HTML]{003300}{($\downarrow$0.8)} & 68.1 \textcolor[HTML]{003300}{($\downarrow$1.2)}\\

{w/o VoteMLLMs}        & 81.0 \textcolor[HTML]{003300}{($\downarrow$1.3)} & 79.2 \textcolor[HTML]{003300}{($\downarrow$0.8)} & 65.3 \textcolor[HTML]{003300}{($\downarrow$4.0)} \\
{w/o OptPrompt}        & 79.0 \textcolor[HTML]{003300}{($\downarrow$3.3)} & 78.6 \textcolor[HTML]{003300}{($\downarrow$1.4)} & 64.9 \textcolor[HTML]{003300}{($\downarrow$4.4)} \\
\arrayrulecolor{lightgray}\midrule
w/o SVD   & 76.8 \textcolor[HTML]{003300}{($\downarrow$5.5)} & 71.6 \textcolor[HTML]{003300}{($\downarrow$8.4)} & 60.5 \textcolor[HTML]{003300}{($\downarrow$8.8)} \\
{SVD$\rightarrow$GPT-4o}         &    79.9  \textcolor[HTML]{003300}{($\downarrow$2.4)}                                  &      78.2  \textcolor[HTML]{003300}{($\downarrow$1.8)}                                 &  68.1  \textcolor[HTML]{003300}{($\downarrow$1.2)} \\
{SVD$\rightarrow$Doubao-V-Pro}         &    \textit{Failed}                                   &         79.1  \textcolor[HTML]{003300}{($\downarrow$0.9)}                              &  69.0 \textcolor[HTML]{003300}{($\downarrow$0.3)}   \\

{SVD$\rightarrow$Qwen3VL}         & \textit{Failed}                                       & \textit{Failed}                                       & 70.0 \textcolor[HTML]{990066}{($\uparrow$0.7)}     \\
\arrayrulecolor{lightgray}\midrule

{w/o DCG}        & 76.3 \textcolor[HTML]{003300}{($\downarrow$6.0)} & 70.9 \textcolor[HTML]{003300}{($\downarrow$9.1)} & 60.1 \textcolor[HTML]{003300}{($\downarrow$9.2)} \\
{Retrieval$\rightarrow$ImgSim} & 78.2 \textcolor[HTML]{003300}{($\downarrow$4.1)} & 77.1 \textcolor[HTML]{003300}{($\downarrow$2.9)} & 64.5 \textcolor[HTML]{003300}{($\downarrow$4.8)}   \\
{Retrieval$\rightarrow$GraphRAG} & 81.1 \textcolor[HTML]{003300}{($\downarrow$1.2)} & 79.5 \textcolor[HTML]{003300}{($\downarrow$0.5)} & 65.5 \textcolor[HTML]{003300}{($\downarrow$3.8)}\\
\arrayrulecolor{black}\bottomrule
\end{tabular}}
\label{tab:ablation_study}
\end{table}

%% file: section/rw.tex
\section{Related Works}



Harmful meme detection has emerged as an important research direction in multimodal semantic analysis.
Previous studies extract visual and textual feature embeddings and fuse them with attention mechanisms~\cite{DBLP:conf/acl-trac/SuryawanshiCAB20, pramanick-etal-2021-momenta-multimodal, lee2021DisMultiHate} and Transformer-based intermediate fusion~\cite{kiela2019supervised, lu2019vilbert, li2019visualbert, chen2020uniter}, and contrastive reweighting approaches~\cite{DBLP:conf/nips/KielaFMGSRT20, muennighoff2020vilio, lippe2020multimodal}. 
With the development of MLLMs, researchers proposed prompt-based~\cite{cao2022prompting, Creative-Harm-23-Ji, cao2023pro, ji2024capalign} and multi-agent-based frameworks for deeper visual understanding~\cite{lin2025ask}. 
Considering the bias in MLLM's inference steps, researchers proposed LLM debate frameworks~\cite{hee2022explaining} to improve the detection performance.
To address challenges in low-resource generalization, researchers have explored few-shot enhanced methods with SFT and RAG methods~\cite{10.1145/3589334.3648145, huang2024towards, DBLP:conf/acl/Liu0LWD25}.
However, harmful memes are ever-shifting on the Internet, but few of these works concerns on improving their generalizability.
Different from these works, {\tool} introduces the explainable design concepts to facilitate the detection of ever-shifting harmful memes.

%% file: section/conclusion.tex
\section{Conclusion}

In this paper, we propose {\tool}, an ever-shifting harmful meme detection method.
We first define DCG, including the harmful types and reproduction steps of how users design harmful memes.
Then, we derive the DCG from historical memes.
Finally, we retrieve and utilize it to guide the MLLMs to detect whether the target meme is harmful.
The evaluation results show that {\tool} can detect harmful memes with 81.1\% accuracy and can be generalized to type-shifting and temporal-evolving memes.
Human evaluation shows that {\tool} can improve the efficiency of human discovery on harmful memes, with 15$\sim$30 seconds per meme.

\section*{Limitations}
Although {\tool} is effective in detecting ever-shifting harmful memes, there are some cases that {\tool} cannot accurately identify their harmfulness.
We manually investigate these bad cases and discuss the reasons for their failures.

First, 60\% of the bad cases come from overly simple meme features. For example, the Minecraft group users always use the cubes to represent humans and objects, but MLLMs have a limited ability to learn these expressions.
In the future, we plan to combine the MLLM fine-tuning and DCG to learn these features.
Second, 35\% of the cases come from the LLM hallucinations. MLLM may violate the guidance of DCG during the inference process and output guessed results, leading to errors in harmful meme detection.
We plan to introduce the stepwise hallucination corrector to revise the correct biases in the responses.
Finally, 5\% of the cases come from newly designed memes with completely unseen design concepts.
The proportion of these memes is small, which has relatively less impact on the detection performance.
We also plan to continuously improve DCG for these cases.

\section*{Ethical Statement}

Our study aims to detect ever-shifting harmful memes on the Internet, but it may raise ethical considerations.
We will discuss how these considerations are mitigated as follows.
First, the dataset may have privacy and license issues because it comes from GOAT-Bench and online Twitter posts.
We control permissions of visitors to the dataset through providing private dataset links and password access, and continuously track the whereabouts of the data to prevent data abuse.
Second, the algorithms may have fairness issues because there are differences in the protection of individual freedom of speech in different countries.
We publicly disclose the DCG and its related guidance, and plan to provide channels for misjudgment on our public website.
Third, the annotators and evaluators may have mental health issues when labeling the dataset.
We have provided regular psychological counseling to annotators, established a job rotation system, and provided reasonable salaries.

\section*{Acknowledgement}

We sincerely appreciate all the reviewers for their
constructive suggestions. This work was supported
by the National Key Research and Development
Program of China (No.2024YFF0618800), National Natural Science Foundation of China Grant No.62402484, No.62232016, Youth Innovation Promotion Association Chinese Academy of Sciences, and Basic Research Program of ISCAS Grant No.ISCAS-JCZD-202405.

%% file: section/appendix.tex
\section{Appendix}
\label{sec:appendix}

\subsection{Detailed Definition of DCG}\label{app:dcg_structure}

\paragraph{Type Node.} 
In this part, we predefined seven macro types for the meme classification, i.e., \textit{Nationality, Gender, Religion, Human, Animal, Culture, and Political}.
Based on this classification system, we can classify the Internet memes into those that represent what the users want to express.

\input{table/type_mapping}

For each macro type, we ask the MLLM to divide it into subtypes to specifically illustrate the details in these memes.
Table \ref{tab:type_mapping} shows the corresponding relationship between macro types and the subtypes. In this table, we only present the example subtypes in DCG's type nodes, where the subtypes will be extended when new memes occur.

With the previous classification system, the macro types and subtypes will be mapped to the type node $\mathcal{N}_\text{type}$'s three-level information.
The prompts $P_T$ for type node generation (Figure \ref{fig:prompt_tree_construction}) will control the subtype extension in the meme reason tree and DCG construction.

\paragraph{Reproduction Nodes.}
In this part, we have defined the design concept, which is the idea and steps that users convey their harmfulness.
We need to reproduce their ideas by observing the visible meme's elements, and the graph represents the MLLM's inner understanding of this meme.

To illustrate the details of the reproduction nodes in the DCG, we have provided an example in Figure \ref{fig:example_dcg}.
In this example, the DCG incorporates four reproduction step nodes and one reproduction goal. The macro type is the "\textit{Gender}" and the subtype is the "\textit{Implicit Misogyny Slangs}".

\paragraph{Representation of Reproduction Steps.}
To represent the graph in the MLLM's input with plain text together with logic-based semantics, we use the logic expression, which is concatenated by logic symbols ($\land, \lor, \lnot, \rightarrow$). The example's reproduction steps can be formulated as follows:
\begin{equation}\label{equa:logic}
\mathcal{N}_{\text{M}_1}\land(\mathcal{N}_{\text{M}_1}\rightarrow\mathcal{N}_{\text{M}_2}\rightarrow\mathcal{N}_{\text{M}_3})\rightarrow\mathcal{N}_{G}  
\end{equation}
where the $\mathcal{N}_\text{rep}^{M_1}$ to $\mathcal{N}_\text{rep}^{M_3}$ means the content in the reproduction nodes, and the $\mathcal{N}_\text{rep}^{G}$ means the content in the reproduction goal.
In the guidance of MLLM's inference of Section \ref{sec:retried}, the input logic expression utilizes the "\textit{then}", "\textit{and}", "\textit{or}", and "\textit{on the contrary}" to replace the logic symbols.
Based on this criterion, the DCG is parsed into text according to the following rules.

\begin{itemize}[leftmargin=*]
\item \textbf{Parsed Text of Reproduction Method:} Directly use the reproduction method's text.
\item \textbf{Parsed Text of Harmful Indicator:} Append the parentheses with labels (i.e., "harmful" and "no harmful") after reproduction methods.
\item \textbf{Parsed Text of Logic Gate and Achievement Edge:} Append the conjunctions (i.e., "\textit{then}", "\textit{and}", "\textit{or}", and "\textit{on the contrary}") between texts of reproduction methods.
\end{itemize}

Therefore, the input design concept with the logic expression can be illustrated as follows:

\begin{center}
\small
\begin{tcolorbox}[colback=white,
                  colframe=black,
                  width=\columnwidth,
                  arc=1mm, auto outer arc,
                  boxrule=0.5pt,
                  left=3pt,
                  right=3pt,
                  top=3pt,
                  bottom=3pt
                 ]
\textbf{\#\# Harmful Type}

Macro Type: Gender; Subtype: Implicit Misogyny Slangs.

\textbf{\#\# Reproduction Step} 

\textit{The malicious user writes a fact text (no harmful), \textbf{then} search the related images based on the fact (no harmful). \textbf{Then}, it specify the image to the biased people (woman, \textcolor{red}{harmful}). The fact \textbf{and} people specification simultaneously cause the attack goal (\textcolor{red}{harmful}): \textcolor{red}{Specify fact to person}, express misogyny to people emotion.}
\end{tcolorbox}
\end{center}


\subsection{Mathematical Rationality for Algorithm \ref{alg:pruning}}\label{app:proof}

\subsubsection{Mathematical Rationality of SVD}\label{app:proof-svd}

We consider a symmetric matrix $\mathbf{A}$ of the form:
\begin{equation}
    \mathbf{A}=\begin{pmatrix}
    \mathbf{A}_\text{Type}&\beta\mathbf{E}_\text{Link}\\
    \beta\mathbf{E}_\text{Link}&\alpha\mathbf{A}_\text{Repr}
    \end{pmatrix}
\end{equation}
where $\mathbf{A}_{\text{Type}}$ and $\mathbf{A}_{\text{Repr}}$ are symmetric matrices representing the adjacency or feature similarity within the type nodes and the reproduction nodes, respectively; $\mathbf{E}_{\text{Link}}$ is the connection matrix between the two parts; and $\alpha$ and $\beta$ are scaling factors satisfying $1 > \alpha > \beta > 0$.

\paragraph{Effect of Scaling Factors.}
To show that the scaling factors $\alpha$ and $\beta$ lead to smaller singular values for the reproduction nodes, we analyze the Rayleigh quotient of $\mathbf{A}$. 
Let $\mathbf{x} = (\mathbf{x}_1, \mathbf{x}_2)$ be a vector partitioned according to the two parts. We can derive the equation:
\begin{equation}
\resizebox{.89\linewidth}{!}{$
    \displaystyle  
\mathbf{x}^{\top}\mathbf{A}\mathbf{x}=\mathbf{x}_1^{\top}\mathbf{A}_\text{Type}\mathbf{x}_1+2\beta\mathbf{x}_1^{\top}\mathbf{E}_\text{Link}\mathbf{x}_1+\alpha\mathbf{x}_2^{\top}\mathbf{A}_\text{Repr}\mathbf{x}_2
$}
\end{equation}

Consider a vector $\mathbf{x}$ that is nonzero only in the reproduction nodes, i.e., $\mathbf{x} = (\mathbf{0}, \mathbf{x}_2)$ with $|\mathbf{x}_2| = 1$, we can derive the equation:
\begin{equation}
\mathbf{x}^{\top}\mathbf{A}\mathbf{x}=\alpha\mathbf{x}_2^{\top}\mathbf{A}_\text{Repr}\mathbf{x}_2\leq\alpha\lambda_\text{max}(\mathbf{A}_\text{Repr})
\end{equation}
where $\lambda_{\max}(\mathbf{A}{\text{Repr}})$ is the largest eigenvalue of $\mathbf{A}{\text{Repr}}$. By the Courant-Fischer minimax theorem, the largest singular value of $\mathbf{A}$ (which equals the largest eigenvalue in absolute value, since $\mathbf{A}$ is symmetric) satisfies:
\begin{equation}
\max_{||\mathbf{x}||=1}|\mathbf{x}^{\top}\mathbf{A}\mathbf{x}|\geq\alpha\lambda_\text{max}(\mathbf{A}_\text{Repr})
\end{equation}

For vectors concentrated on the reproduction nodes, the Rayleigh quotient scales with $\alpha$, i.e., the singular values associated with the reproduction nodes are roughly proportional to $\alpha$. Similarly, for vectors that involve both parts, the Rayleigh quotient scales with $\beta$.
Since $\alpha < 1$ and $\beta < 1$, the scaling reduces the Rayleigh quotient for vectors involving the reproduction nodes, leading to \textbf{smaller singular values} for those modes.

\paragraph{Low-rank Approximation and Pruning.}
In the SVD-based pruning, we approximate $\mathbf{A}$ by a low-rank matrix $\tilde{\mathbf{A}}_k = \mathbf{U}_k \boldsymbol{\Sigma}_k \mathbf{V}_k^\top$, where only the $k$ largest singular values are retained. Since the scaling factors $\alpha$ and $\beta$ decrease the singular values corresponding to the reproduction nodes, these values are more likely to be truncated in the low-rank approximation. Consequently, the edges within the parts are identified as redundant and can be pruned.

Thus, the scaling strategy with $\alpha$ and $\beta$ is mathematically justified as it systematically reduces the influence of the reproduction nodes in the singular value spectrum, enabling reproduction nodes' pruning while preserving the type nodes.

\subsubsection{Mathematical Rationality for Logarithmic Cut-off Determination}\label{app:proof-cutoff}

\paragraph{Properties of Long-Tail Distributions.} Let $\{x_i\}_{i=1}^n$ be a set of $n$ positive real-valued observations assumed to have a long-tail distribution. After sorting in descending order, we denote: $x_{1} \geq x_{2} \geq ... \geq x_{n} > 0$.
Our target is to find a threshold $x_c$ such that:
\begin{itemize}[leftmargin=*]
    \item $x_i \geq x_c$ belongs to the "head" region.
    \item $x_i < x_c$ belongs to the "tail" region.
\end{itemize}

The long-tail distributions exhibit the fundamental property that their generating processes are \textit{multiplicative} rather than additive. This leads to the following characterization:

\begin{theorem}[Logarithmic Normalization]
For a random variable $X$ following a long-tail distribution, the logarithmic transform $Y = \ln(X)$ typically follows a distribution with exponentially decaying tails (e.g., normal distribution for log-normal, exponential for power-law).
\end{theorem}

This theorem converts multiplicative differences in the original scale to \textbf{additive} differences in the \textbf{logarithmic} scale.

\paragraph{Extreme Value Behavior in Logarithmic Space.}

Let $Y_{1} \geq Y_{2} \geq ... \geq Y_{n}$ be the order statistics of the log-transformed data with a long-tail distribution. 
The spacings between consecutive order statistics satisfy:

\begin{lemma}[Spacing Distribution]
For observations from a distribution with continuous density $f_Y(y)$, the normalized spacings 
$n \cdot (Y_{i} - Y_{i+1}) \cdot f_Y(Y_{i})$ converge to independent standard exponential random variables as $n \to \infty$.
\end{lemma}

This lemma suggests that in regions where the density $f_Y(y)$ is relatively constant, the gaps $\Delta_i$ should be approximately equally sized. At the boundary between the head and tail regions, we expect a \textbf{discontinuity} in the density, leading to an \textbf{anomalously large gap}.

\paragraph{Optimality of Maximum Gap Selection.}
We provide the theorem for maximum gap optimality:

\begin{theorem}[Maximum Gap Optimality]
Assume the data generation follows two regimes:
\begin{itemize}[leftmargin=*]
    \item \textbf{Head regime:} $Y \sim F_1$ with support on $[a, \infty)$
    \item \textbf{Tail regime:} $Y \sim F_2$ with support on $(-\infty, b]$
    \item With $a > b$ and minimal overlap between regimes
\end{itemize}

As $n \to \infty$, the probability that the maximum gap occurs at the true boundary converges to 1.
\end{theorem}

Then, we provide the proof for this theorem:

\begin{proof}
Let $Y_{k^*}$ and $Y_{k^*+1}$ be the order statistics straddling the true boundary. The expected gap at this position is:
\begin{equation}
    \mathbb{E}[\Delta_{k^*}] = \mathbb{E}[Y_{k^*} - Y_{k^*+1}] \geq a - b + o(1)
\end{equation}
whereas for position $i \neq k^*$, the expected gap is:
\[
\mathbb{E}[\Delta_i] = O\left(\frac{1}{n \cdot f_Y(Y_{i})}\right)
\]
Since $a - b > 0$ represents a fixed separation between regimes, and the other gaps shrink as $O(1/n)$, we have:
\[
\lim_{n \to \infty} \mathbb{P}\left(\Delta_{k^*} > \max_{i \neq k^*} \Delta_i\right) = 1
\]
Thus, the maximum gap asymptotically identifies the true boundary.
\end{proof}



\subsection{Case Study of DCG Construction and Harmful Meme Detection}

\begin{figure*}[t]
\centering
\includegraphics[width=\textwidth]{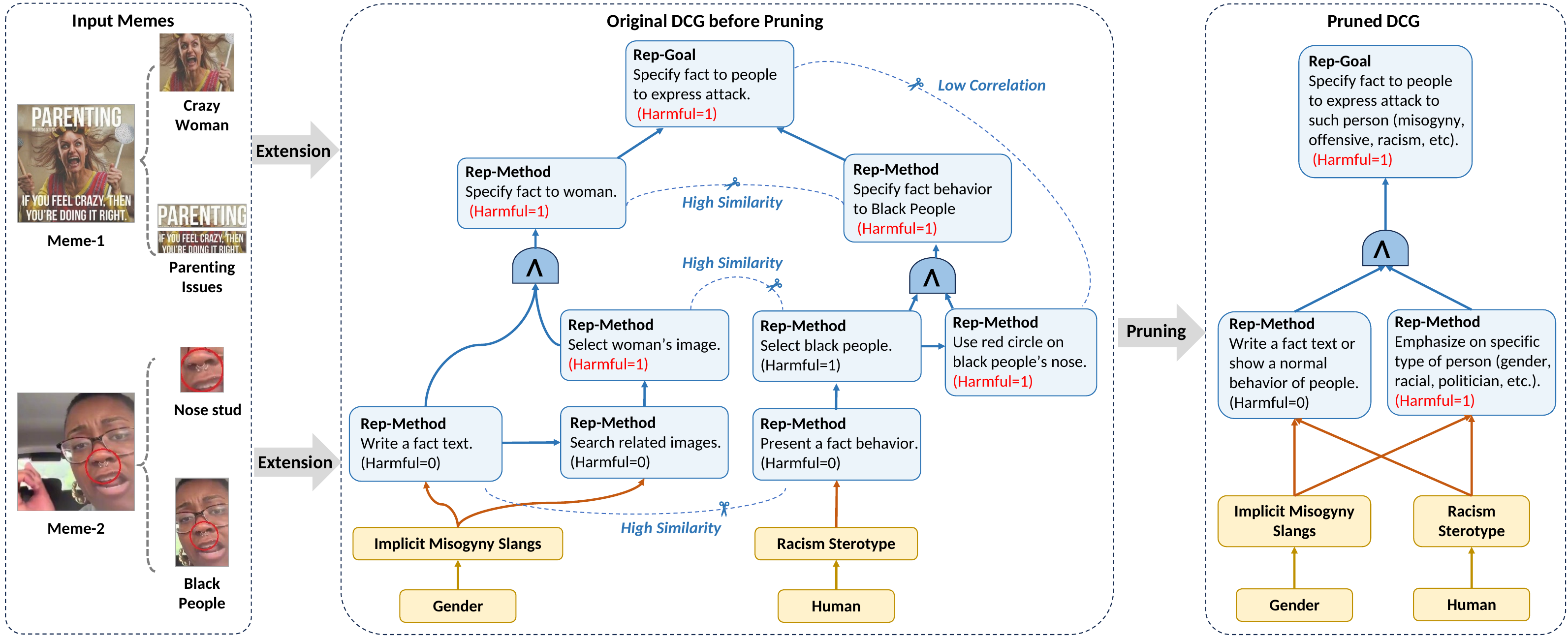}
\caption{The case of {\tool}'s DCG pruning.}
\label{fig:dcg_pruning_case}
\end{figure*}

In this section, we illustrate the case of how the DCG is pruned in the {\tool}'s second step, including the node extension, score calculation in the SVD, and the details are shown in Figure \ref{fig:dcg_pruning_case}.

In this case, we can see that {\tool} first analyzes the surface-level elements of the two historical memes from Figure \ref{fig:motivation}'s motivation example.
The malicious users incorporate the "\textit{crazy woman}", "\textit{parenting issues}", "\textit{nose stud}", and "\textit{black people}" in these two memes, which are apparently shown in the elements.

Then, {\tool} extends the elements to the inner ideas, which are the reproduction steps of the design concept that reflect what users have done to design such harmful memes. 
The two memes are extended to the DCG with 13 nodes (nine reproduction step nodes and four type nodes).
From bottom to top, the DCG represents what the users have done to express misogyny and racism.
However, the graph contains too many nodes and edges, which makes it difficult to understand.

To reduce the DCG's scale, we prune it based on the SVD, which is calculated based on the node's similarity and root correlation.
For the node's similarity, we find that the reproduction methods like "\textit{select woman's image}" and "\textit{select black people}" have high TF-IDF similarity, where "\textit{circle the black people's nose}" and the "\textit{specify fact to people}" seem to have low correlation that need to be calibrated and pruned.
After SVD pruning and merging, these edges and nodes are removed in the DCG, which only keeps the core part that has seven nodes to represent the user's design concept.

\subsection{Other Experimental Results}\label{app:other_res}

\paragraph{Pruning Time Cost of SVD Replacement.}

To evaluate the time cost between SVD and MLLMs for graph pruning, we separately evaluate the time the {\tool} takes when pruning the historical memes $\{M_1^{H},...,M_n^{H}\}$. The time cost is calculated with the following equation:
\begin{equation}
Cost_\text{time}=Sec(\mathcal{G}_R\rightarrow\mathcal{G}^{'}_R)/N(\{M_i^{H}\})
\end{equation}
where the DCG $\mathcal{G}_R$ is prepared based on the historical meme $M_i^{H}$'s analysis results.
With this equation, we do not consider the time cost of graph generation, but only analyze how effectively the graph can be pruned.
The tradeoff between the model performance and the efficiency illustrates the benefit of {\tool}.

\input{table/tab_time_cost}

Table \ref{tab:prune_time_cost} illustrates the pruning time cost of the SVD and the other MLLMs. We can see that, MLLMs can prune the DCG, but they have too much time cost, which is over 10$\times$ to the SVD.
In practice, these methods are not applicable, which may come from the fact that the input graph has too many tokens, as well as the long CoT and inference steps that reduce their efficiency.
However, we can also see that Qwen3VL-235B can slightly improve the detection performances in the TE task, which means if we can use SVD's results as ground-truth and fine-tune these MLLMs to an acceptable time range (around 2$\times$ to 5$\times$ to SVD), the MLLMs may be useful to improve {\tool}'s performance.

\paragraph{Hyperparameter Analysis.}
We also evaluate the performance of {\tool} when we tune the Retained Nodes' Proportion $\theta$ from $0\%\sim100\%$ with a 5\% interval.
We evaluate the performance of GPT-4o and Doubao-1.5-Vision-Pro on these intervals, where we calculate the \textbf{average results} of the OOD and TE experiment.

\begin{figure}[t]
\centering
\includegraphics[width=\columnwidth]{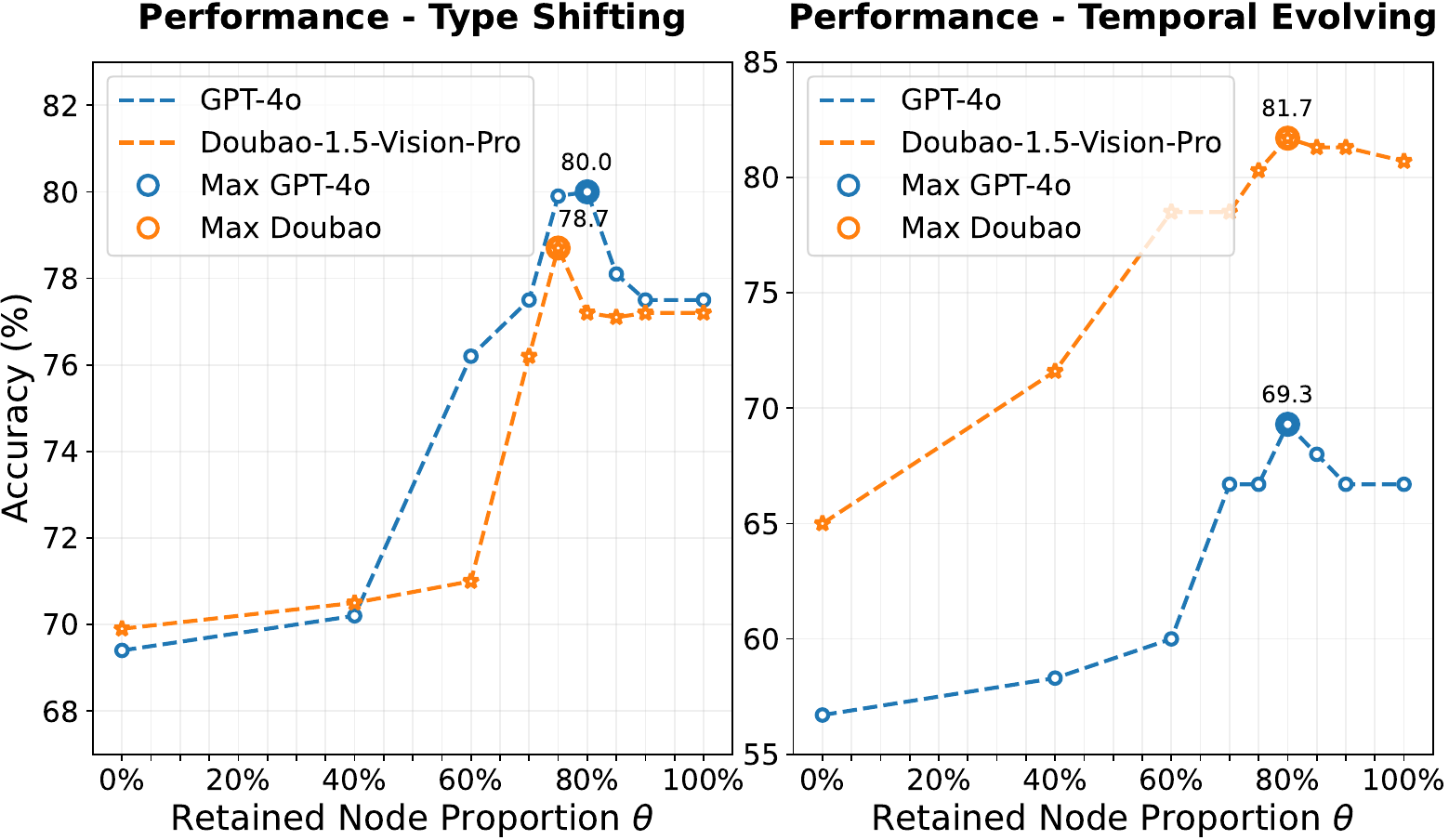}
\caption{The effect of the retained node's proportion $\theta$.}
\label{fig:parameter_proportion}
\end{figure}

\begin{figure}[t]
\centering
\includegraphics[width=\columnwidth]{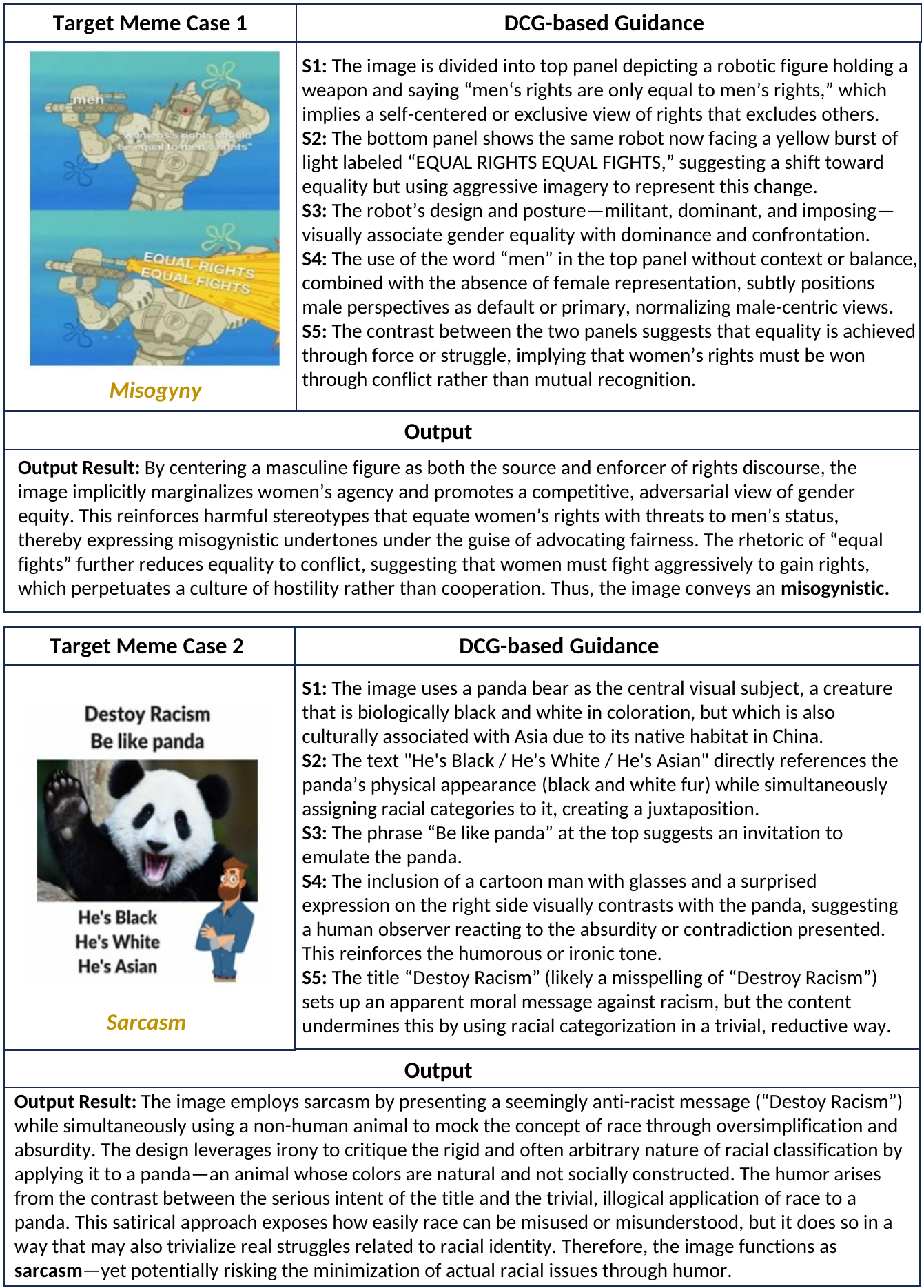}
\caption{The case study of {\tool} of other two meme's Qwen2.5VL's outputs.}
\label{fig:tool_case_study}
\end{figure}

Figure \ref{fig:parameter_proportion} shows the results of hyperparameter tuning.
We can see that, most of the highest value comes from $\theta=75\%$ and $80\%$, where the difference between these two parameters is slightly different.
We also manually check the values across different types and temporal quarters, and find that the highest results all distribute in the proportion within $70\%\sim80\%$.
Therefore, we choose the balance value for all the experiments, i.e., 75\%, for the {\tool}'s SVD pruning.

\subsection{Case Study of Other Harmful Memes}

In Figure \ref{fig:tool_case_study}, we provide two more memes' Qwen2.5VL's output based on our DCG guidance. The output can correct the misleading analysis (e.g., the original racism prediction in case 2) from these memes and output the correct harmful types (e.g., sarcasm of case 2), illustrating its usefulness in harmful meme detection.

\subsection{Questionnaire for Human Evaluation}\label{app:human}

We have provided the content of the questionnaire provided to evaluators, including the introduction of the 5-point Likert scale, the detailed definition for the five criteria, as well as the core dimensions for evaluators to rate our DCG and memes.
We also suggest evaluators provide open feedback to our DCG's improvement, so {\tool} can be continuously optimized.

\input{algo/questionnaire}



\subsection{Safeguarding the Text-to-Image Models}\label{app:text-to-image-safeguard}

After the previous human evaluation that illustrates the benefits of {\tool}, we think further on the usefulness of {\tool}, i.e., whether the DCG can cheat the Text-to-Image models and output harmful information.
Theoretically, if one MLLM cannot correctly detect the harmfulness of the meme, it may be misled by these design concepts without security alignment.
Inspired by it, if we can successfully jailbreak the MLLMs to generate such harmful memes, we may have the chance to protect them based on the design concepts:

\textbf{Step-1: Attack Prompt Construction:} We obtain the target harmful meme $M^{T}_i$ and its stepwise guidance, then we directly form it as the prompt that will be input to the MLLM:

\begin{center}
\small
\begin{tcolorbox}[colback=white,
                  colframe=black,
                  width=\columnwidth,
                  arc=1mm, auto outer arc,
                  boxrule=0.5pt,
                  left=3pt,
                  right=3pt,
                  top=3pt,
                  bottom=3pt
                 ]
\textbf{\#\# Task of Qwen-Image} 

Please draw a meme graph with humor.
              
\textbf{\#\# Reproduction Step} 

\textit{The malicious user writes a fact text sentence like "parenting is too difficult and makes people crazy", then draws a crazy lady in the image.}
\end{tcolorbox}
\end{center}

\textbf{Step-2: DCG-Guard Safeguarding:} We ask another Qwen3VL-235B model to refer to the DCG, then safeguard the prompt and remove the biased information in the memes, and add the forbidden sentences. The sentences like "\textit{crazy lady in the image}" will be replaced with commands like "\textit{\textbf{stop generating such image because of misogyny}}", or replace them with other benign elements like "\textit{\textbf{parents that are raising children}}".

\input{table/safeguarding}

\begin{figure*}[t]
\centering
\includegraphics[width=\textwidth]{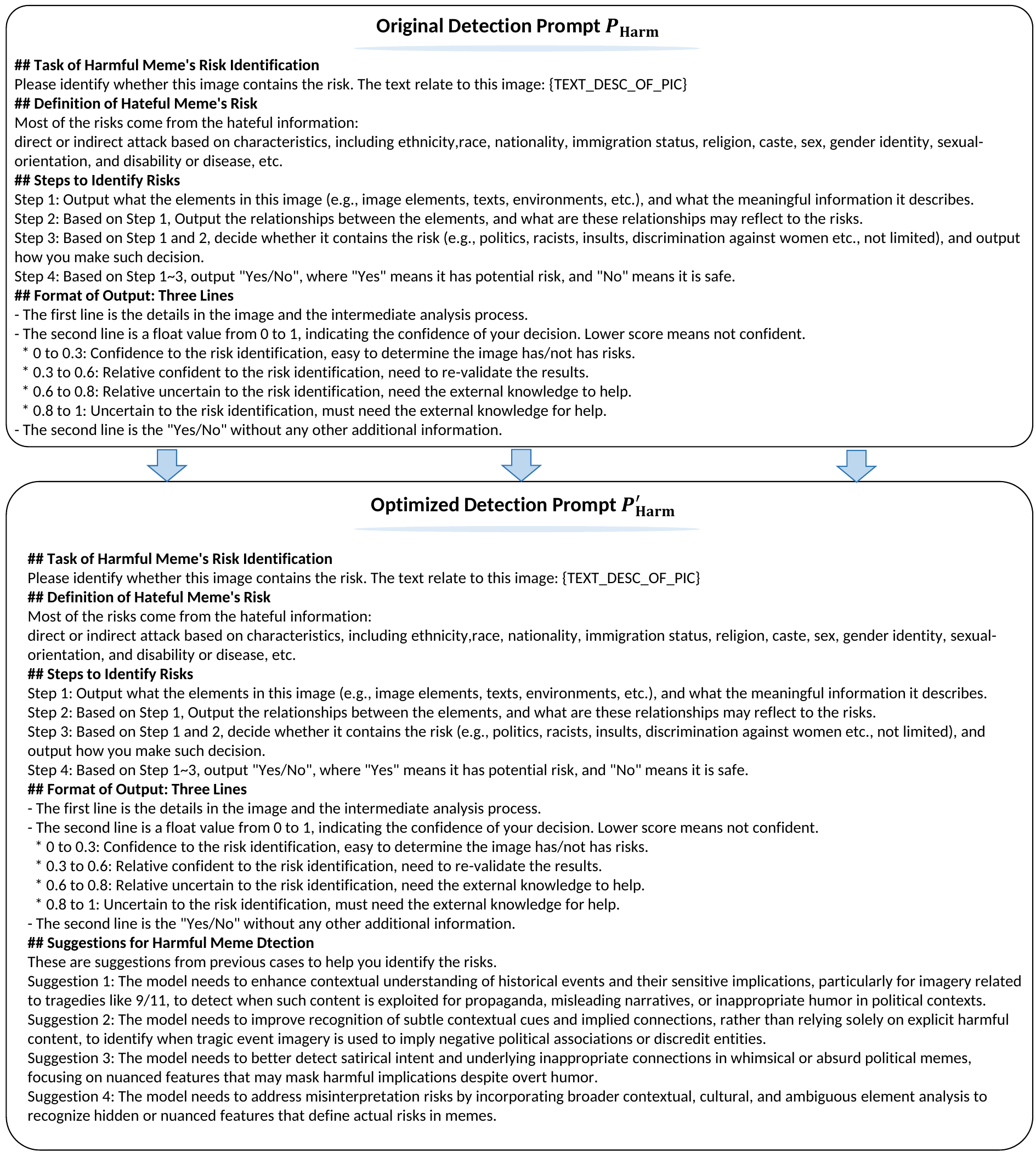}
\caption{The detection prompt $P_\text{Harm}$ and $P^{'}_\text{Harm}$.}
\label{fig:detection_prompt}
\end{figure*}

\begin{figure*}[t]
\centering
\includegraphics[width=\textwidth]{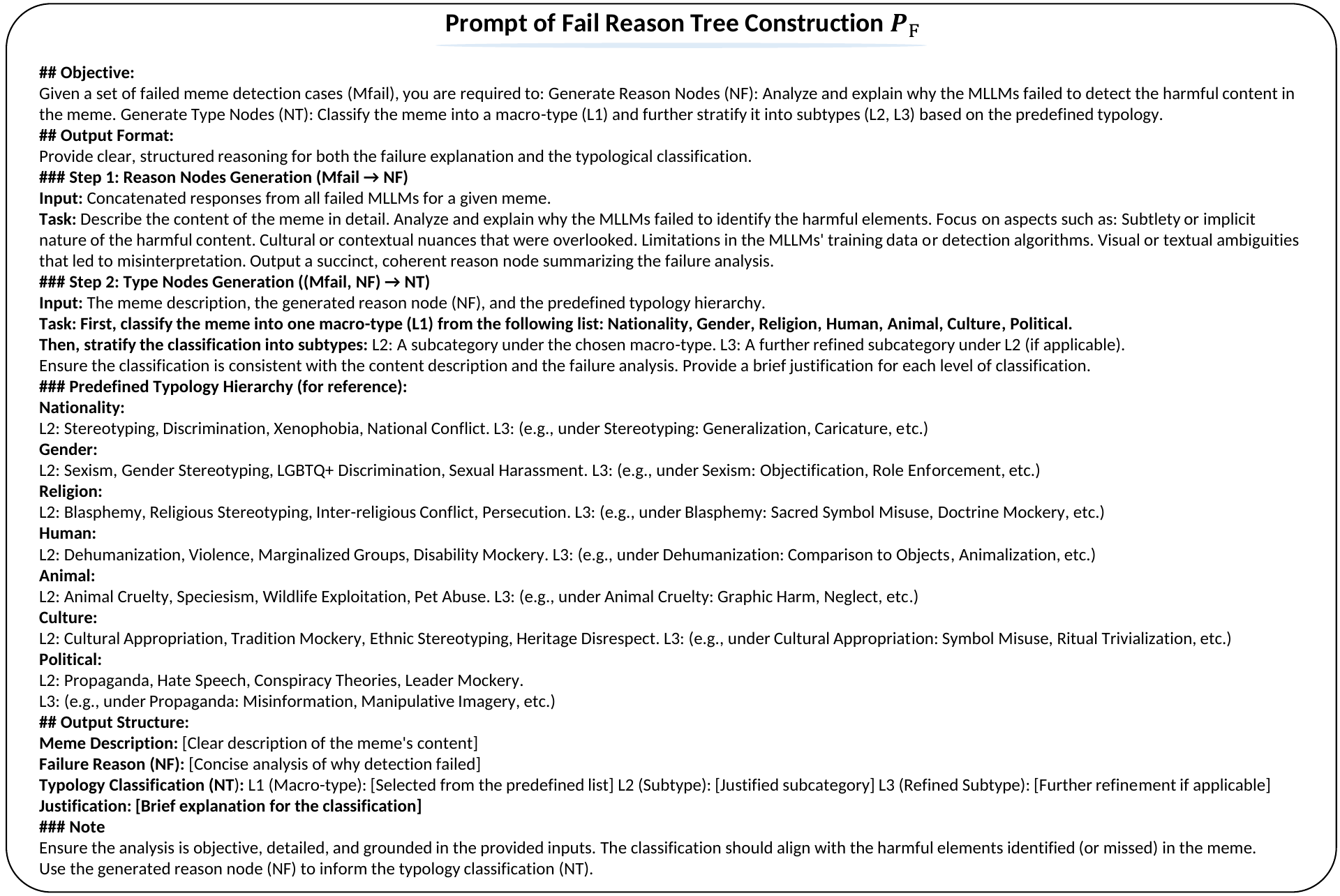}
\vspace{-0.6cm}
\caption{The fail reason tree's construction prompt $P_\text{F}$.}
\label{fig:prompt_tree_construction}
\vspace{-0.6cm}
\end{figure*}

\begin{figure*}[t]
\centering
\includegraphics[width=\textwidth]{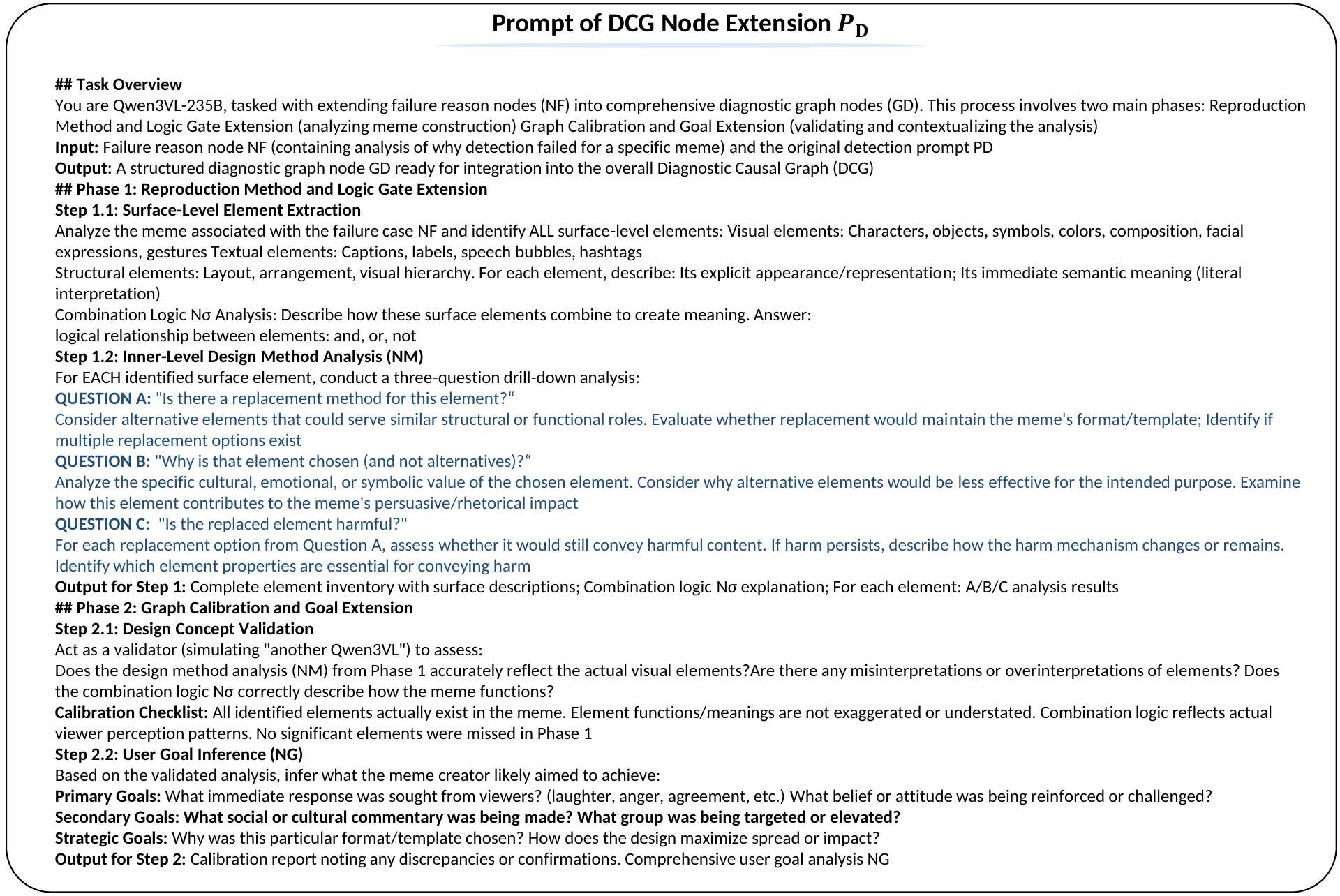}
\vspace{-0.6cm}
\caption{The DCG construction's prompt $P_\text{D}$.}
\label{fig:prompt_dcg_construction}
\vspace{-0.6cm}
\end{figure*}

We select the Qwen-Image as the target MLLM, which is a model that has not been successfully aligned and may generate harmful meme content, and randomly choose 10$\sim$12 cases as the target memes.
We introduce two parts of the evaluation metrics. First, we evaluate how many images are successfully generated based on our attack prompt, and calculate the success rates of the attack prompts. Then, we evaluate the average
\textbf{SSIM}, which is the metric that represents the similarity between AI-reconstructed images and original images.
We evaluate the result with/without DCG-Guard on the five harmful types of memes.

Table \ref{tab:dcg_safeguard} shows the results of safeguarding the harmful images through DCG-Guard. We can see that, before safeguarding the Qwen-Image, some attack prompts can achieve nearly 100\% success rates (except for the Toxicity memes that contain apparent harmful features in the prompt).
With the help of DCG-Guard, the success rates reduce by over 50\%, where most of the harmful information in memes is detected and refused to generate with the stop commands.
Moreover, even if some images are successfully generated, they may be different from the previous images based on the benign element replacement, with over 0.3 SSIM decrease.
These results show that DCG can also help safeguard text-to-image models.

\subsection{Prompts in {\tool}}\label{app:prompt_details}

In this section, we show all the details of the prompts that we have used in {\tool}

\begin{itemize}[leftmargin=*]
    \item \textbf{Harmful meme detection:} $P_\text{Harm}, P^{'}_\text{Harm}$ in Figure \ref{fig:detection_prompt}, where $P_\text{Harm}$ is the original prompt before generating the target guidance (the Vanilla MLLMs use this prompt), and $P^{'}_\text{Harm}$ is the prompt after generating the target guidance.
    \item \textbf{Fail reason tree construction:} $P_\text{F}$ in Figure \ref{fig:prompt_tree_construction}, which indicates the steps of how to find the fail cases and generate the reasons.
    \item \textbf{DCG construction:} $P_\text{D}$ in Figure \ref{fig:prompt_dcg_construction}, which indicates how to generate and prune the nodes/edges in DCGs based on the basic structure.
\end{itemize}










%% file: table/type_mapping.tex
\begin{table}[t]
\small
\caption{The macro type and example subtypes.}
\vspace{-0.2cm}

\resizebox{\columnwidth}{!}{
\begin{tabular}{c|l}
\toprule

\textbf{Macro-Type}          & \multicolumn{1}{c}{\textbf{Example Subtypes in DCG's Type Part}} \\
\midrule
\multirow{3}{*}{\textbf{Nationality}} & Countryhumans                                 \\
                             & Historical Event Parody                        \\
                             & National Stereotype                           \\
                             \arrayrulecolor{lightgray}\hline
\multirow{4}{*}{\textbf{Gender}}      & Gender Role Reversal                          \\
                             & Implicit Misogyny Slangs                      \\
                             & Performative-Male Symbols                     \\
                             & Transgender Symbols                           \\
                             \arrayrulecolor{lightgray}\hline
\multirow{4}{*}{\textbf{Religion}}    & Holiday \& Ritual Meme Templates               \\
                             & Scripture \& Figure Parody                     \\
                             & Islamic Muslim Symbols                        \\
                             & Buddhist Symbols                              \\
                             \arrayrulecolor{lightgray}\hline
\multirow{2}{*}{\textbf{Human}}       & Racism Sterotype                              \\
                             & Disability Stereotype                         \\
                             \arrayrulecolor{lightgray}\hline
\multirow{3}{*}{\textbf{Animal}}      & Personified Animal Memes                      \\
                             & Animal Behavior as Metaphor                   \\
                             & Mimicry in Nature as Meme Template            \\
                             \arrayrulecolor{lightgray}\hline
\multirow{4}{*}{\textbf{Culture}}     & Nonsense Literature                           \\
                             & Abbreviation Culture                          \\
                             & Versailles Literature                         \\
                             & Video Game                                    \\
                             \arrayrulecolor{lightgray}\hline
\multirow{3}{*}{\textbf{Political}}   & Politician as Meme Template                   \\
                             & Political Figure's Biological Reduction       \\
                             & Policy \& Event Satire                        \\
                            \arrayrulecolor{black} \bottomrule
\end{tabular}}
\label{tab:type_mapping}
\end{table}

%% file: table/tab_time_cost.tex
\begin{table}[htbp]
\vspace{-0.3cm}
\caption{The pruning time cost of variants in SVD replacement (seconds per meme for DCG pruning).}
\vspace{-0.2cm}
\resizebox{\columnwidth}{!}{
\begin{tabular}{l|lll}
\toprule
\textbf{Variants} & \textbf{ID}       & \textbf{OOD}       & \textbf{TE}        \\
\midrule
RepMD w/ SVD      & 4.9               & 6.1                & 5.3                \\
SVD$\rightarrow$GPT-4o       & 62.3 ($13\times$) & 61.3 ($10\times$) & 52.9 ($10\times$) \\
SVD$\rightarrow$Doubao-V-Pro & \textit{Failed}            & 71.2 ($12\times$) & 69.7 ($13\times$) \\
SVD$\rightarrow$Qwen3VL      & \textit{Failed}            & \textit{Failed}             & 75.5 ($14\times$)\\
\bottomrule
\end{tabular}}
\vspace{-0.4cm}
\label{tab:prune_time_cost}
\end{table}

%% file: algo/questionnaire.tex
\begin{center}
\small
\begin{tcolorbox}[colback=white,
                  colframe=black,
                  width=\columnwidth,
                  arc=1mm, auto outer arc,
                  boxrule=0.5pt,
                  left=3pt,
                  right=3pt,
                  top=3pt,
                  bottom=3pt
                 ]

\textbf{Questionnaire: Evaluation of Design Concept Graph (DCG) for Harmful Meme Detection}

\vspace{0.2cm}

\textbf{Instructions:}  
Please evaluate the provided \textbf{Design Concept Graph (DCG)} in relation to the corresponding \textbf{meme image} based on the following criteria.  
For each statement, select the response that best matches your assessment using the \textbf{5-point Likert scale}, where:

\noindent\textbf{1 = Strongly Disagree}  

\noindent\textbf{2 = Disagree}  

\noindent\textbf{3 = Neutral (neither good nor bad / borderline)}

\noindent\textbf{4 = Agree}  

\noindent\textbf{5 = Strongly Agree}

\textbf{Note:} A score of \textbf{3 represents the threshold between positive and negative evaluation}. Scores above 3 indicate a favorable assessment, while scores below 3 indicate an unfavorable assessment.

\vspace{0.2cm}

\textbf{Part 1: Criteria Definition}

\noindent\textbf{(1) Relevance (R):} The DCG is closely related to the content and meaning of the meme image.

\noindent\textbf{(2) Correctness (C):} The DCG accurately represents the key elements and message of the meme image.

\noindent\textbf{(3) Actionability (A):} The DCG provides clear and practical insights that could guide further actions (e.g., content moderation, design adjustments, or analysis).

\noindent\textbf{(4) Uniqueness (U):} The DCG is distinct and differs meaningfully from graphs of other similar meme images.

\noindent\textbf{(5) Explainability (E):} The DCG effectively explains how the harmful meme image was designed and reflects the creator's intent or thought process.

\vspace{0.2cm}

\textbf{Part 2: Core Dimensions}

\vspace{0.1cm}

\begin{tabular}{lll}
\textbf{No.} & \textbf{Dimension} & \textbf{Rating (1--5)} \\
\hline
1 & Relevance (R) & 1 2 3 4 5 \\
2 & Correctness (C) & 1 2 3 4 5 \\
3 & Actionability (A) & 1 2 3 4 5 \\
4 & Uniqueness (U) & 1 2 3 4 5 \\
5 & Explainability (E) & 1 2 3 4 5 \\
\end{tabular}

\vspace{0.2cm}

\textbf{Part 3: Open Feedback (Optional)}  
Please provide any additional comments or suggestions regarding the DCG or its evaluation:

\rule{\textwidth}{0.5pt}

\begin{center}
\textbf{Thank you for your participation!}
\end{center}

\end{tcolorbox}
\end{center}

%% file: table/safeguarding.tex
\begin{table}[t]
\small
\caption{The safeguarding of harmful image generation.}
\vspace{-0.2cm}

\resizebox{\columnwidth}{!}{
\begin{tabular}{l|ll|l}
\toprule
\multicolumn{1}{c|}{\multirow{2}{*}{\textbf{Metric}}} & \multicolumn{2}{c|}{\textbf{Generated Img. / Total Guidance}} & \multicolumn{1}{c}{\multirow{2}{*}{{$\mathbf{\Delta}$}}} \\
\multicolumn{1}{c|}{}                                 & w/o DCG-Guard            & w/ DCG-Guard                   & \multicolumn{1}{c}{}                                \\
\midrule
Racism                                               & 7/12 (58.3\%)             & 1/12 (8.3\%)                    & 50.0\%                                               \\
Misogyny                                             & 8/10 (80.0\%)             & 1/10 (10.0\%)                   & 70.0\%                                               \\
Offensiveness                                        & 7/10 (70.0\%)             & 1/10 (10.0\%)                   & 60.0\%                                               \\
Sarcasm                                              & 10/10 (100.0\%)           & 0/10 (0.0\%)                    & 100.0\%                                               \\
Toxity                                               & 1/10 (10.0\%)             & 0/10 (0.0\%)                    & 10.0\%                                               \\
\midrule
\multicolumn{1}{c|}{\multirow{2}{*}{\textbf{Metric}}} & \multicolumn{2}{c|}{\textbf{Average SSIM}}                    & \multicolumn{1}{c}{\multirow{2}{*}{$\mathbf{\Delta}$}} \\
\multicolumn{1}{c|}{}                                 & w/o DCG-Guard            & w/ DCG-Guard                   & \multicolumn{1}{c}{}                                \\
\midrule
Racism                                               & 0.76                     & 0.12                           & 0.64                                                \\
Misogyny                                             & 0.90                     & 0.33                           & 0.57                                                \\
Offensiveness                                        & 0.54                     & 0.19                           & 0.35                                                \\
Sarcasm                                              & 0.82                     & {-}          & {-}                               \\
Toxity                                               & 0.24                     & {-}          & {-}    \\
\bottomrule
\end{tabular}}
\vspace{-0.3cm}
\label{tab:dcg_safeguard}
\end{table}